\documentclass[lettersize,journal]{IEEEtran}
\usepackage{amsmath,amsfonts}
\usepackage{algorithmic}
\usepackage{algorithm}
\usepackage{array}
\usepackage[caption=false,font=normalsize,labelfont=sf,textfont=sf]{subfig}
\usepackage{textcomp}
\usepackage{stfloats}
\usepackage{url}
\usepackage{verbatim}
\usepackage{graphicx}
\usepackage{cite}
\usepackage{graphicx} 
\usepackage{rotating}
\usepackage{adjustbox}
\usepackage{xcolor}
\begin{document}

\title{Confined Space Underwater Positioning Using Collaborative Robots}

\author{

Xueliang Cheng~(\IEEEmembership{Student Member,~IEEE,}), Kanzhong Yao~(\IEEEmembership{Student Member,~IEEE,}), \\ Andrew West, Simon Watson~(\IEEEmembership{Member,~IEEE,}), Ognjen Marjanovic~(\IEEEmembership{Member,~IEEE,}), \\ Barry Lennox~(\IEEEmembership{Senior Member,~IEEE,}) and Keir Groves~(\IEEEmembership{Member,~IEEE})
        % <-this % stops a space
% \thanks{This paper was produced by the IEEE Publication Technology Group. They are in Piscataway, NJ.}
% <-this % stops a space
% \thanks{Manuscript received XXXX XX, 2025; revised XXXX XX, 2025.}

}

\markboth{Journal of \LaTeX\ Class Files,~Vol.~XX, No.XX,XX XXXX}%
{Shell \MakeLowercase{\textit{et al.}}: A Sample Article Using IEEEtran.cls for IEEE Journals}


\maketitle

\begin{abstract}
Positioning of underwater robots in congested and enclosed spaces remains unsolved for field operations.
The existing field ready systems are generally suited to use in large, open marine environments.
In enclosed and congested environments, which are common in industrial settings, existing systems suffer from a mixture of issues, such as: poor coverage, reliance on added infrastructure and the need for feature rich environments.
\textcolor{green}{
Acoustic-based positioning, commonly used in marine environments, faces challenges in industrial underwater settings. Multipath effects from continuous sound reflections and interactions with boundaries can increase signal noise, reducing positioning accuracy and reliability.
}
Accurate and readily deployable positioning is a prerequisite for performing repeatable autonomous missions and therefore, until now, there has been a technological bottleneck restricting robotic deployments.
The Collaborative Aquatic Positioning system presented in this paper uses a mixture of collaborative robotics and sensor fusion to solve the problem.
\textcolor{green}{A unique aspect of this system is inspired by the concept of a mother-ship, where the surface vehicle acts as a "leader" to assist in the positioning of the underwater robot. This innovative approach, particularly in the context of a free-moving surface vehicle, enables positioning even in GPS-denied and highly constrained environments.}
The proposed positioning system is deployed in a large water tank and repeatable autonomous missions are performed using the system's position measurement for real-time trajectory control.
Experimental results show that the system can achieve a mean Euclidean distance (MED) error of 70~mm while operating in real-time.
The system enables almost complete coverage of the body of water in large pools without requiring fixed infrastructure, lengthy calibration, or feature rich environments.
The Collaborative Aquatic Positioning system builds upon recent advances in mobile robot sensing and a recently developed leader follower control system to provide a step-change in positioning capability for real-world, high-precision autonomous underwater navigation. 
\end{abstract}


%%%%%%%%%%%%%%%%%% CHAPTER 1 %%%%%%%%%%%%%%%%%%
\section{Inroduction}\label{sec:intro} % probably better to use \input{} for each chapter to draw content from other files rather than having everything in one big file
Over the last decade, the field of underwater robotics has grown substantially.
Today, the deployment of Remotely Operated Vehicles (ROVs) is both safe and routine, extending beyond offshore industries~\cite{reachrobotics} to include operations in spatially restricted aquatic environments.
Presently, the absence of an adequate positioning system represents a significant technological bottleneck, hindering the introduction of higher levels of autonomy in the field use of underwater robots in industrial scenarios.

Thus far there has not been an underwater positioning system available, with sufficient accuracy, scalability and real world viability, to facilitate the use of autonomous robotics in typical industrial underwater settings.
The goal of this ongoing research is to provide an underwater positioning system that can function in highly physically constrained environments~\cite{4089043,ayoola2019situ} with sufficient accuracy to facilitate repeatable and reliable autonomous robotic missions in real-world scenarios~\cite{1234567}.
To maximize applicability in real-world scenarios, the positioning system should operate without reliance on fixed infrastructure (e.g., cameras), be able to cover extensive areas with minimal blind spots, and maintain positional errors (MED) below approximately 100 mm to enable navigation in cluttered environments.

\subsection{Highly constrained underwater environments}

\begin{figure}[htbp]
    \centering
    \includegraphics[width=\linewidth]{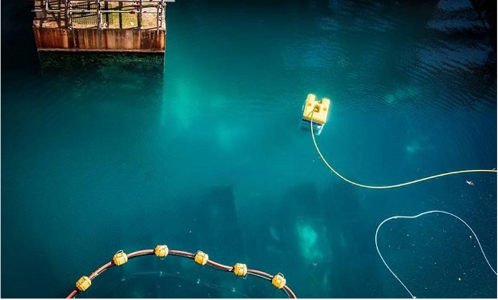}
    \caption{The Saab Seaeye Tiger has spent five years working in nuclear storage ponds~\cite{sella}}
    \label{fig:sella}
\end{figure}

There are numerous categories of highly physically constrained underwater environments that must be accessed on a regular basis for purposes such as inspection, maintenance, repair, and decommissioning.
These environments are characterized by limited spatial conditions, which may include confinement by walls, narrow passageways, or areas that are densely populated with obstacles.
Examples of such environments include nuclear fuel storage pools~\cite{griffiths2016avexis} (Figure~\ref{fig:sella}), liquid storage facilities~\cite{duecker2019learning}, flooded mines~\cite{alvarez2018underwater}, ship hulls~\cite{song2020review} and pipelines~\cite{zhao2022offshore}.

%
% In many cases, regular inspection of these environments is a regulatory necessity.
% %
% For example, the Harmonized System of Survey and Certification (HSSC) Guidelines, Resolution A.1140~\cite{hssc_guidelines_resolution_a114031}, specify that the outside of a passenger ship's hull should be carried out twice in a five year period.

Traditionally, accessing highly constrained underwater environments requires human divers.
However, working in these environments can involve high levels of risk, be tedious and also expensive~\cite{lin2023ship}.
This reality, alongside technological advancements and the need to reduce costs, has led to increased use of robotic vehicles to access such areas.
In addition, small robots can be used to operate in environments where human access would not be possible due to physical constraints, for example, access through pipes or openings that are too narrow for humans~\cite{fackler2017six}.

\subsection{Positioning for underwater robots}
Typically robotic underwater vehicles are split into two categories, Remote Operated Vehicles (ROVs) and Autonomous Underwater Vehicles (AUVs).
Whilst ROVs are remotely driven by a human operator, AUVs are autonomous vehicles that receive high-level commands from the operator, such as a list of waypoints that the robot must navigate to.
AUVs typically necessitate a positioning system to facilitate autonomous navigation.
Conversely, remotely operated vehicles (ROVs) do not inherently require such systems for operation; however, incorporating a positioning system can enhance their performance.

For remotely driven ROV missions, there are two main benefits associated with the provision of accurate positioning.
First, the operator has an additional source of information, which can aid navigation.
This assists the operator in driving through the environment and not losing track of the robot's location, which is a common issue~\cite{li2021command}.
Second, an accurate positioning system allows any data from the sensor payload to be spatially located (geo-tagged), meaning that sensor readings can be repeatably mapped in the underwater environment and presented in human readable formats, such as a heat map format.

For AUVs, an accurate positioning system is essential~\cite{palomeras2019autonomous}, with effective autonomous navigation relying on regular, accurate position updates.
There are several well documented benefits to performing fully autonomous unmanned robotic missions, where the operator has minimal input. These include cost reduction, improved repeatability, increased survey frequency.
Aside from fully autonomous systems, lower levels of autonomy, such as position and velocity control, which can provide smooth and accurate navigation in the presence of disturbances, also require accurate positioning information.
%
% Safety concerns and cost-effectiveness are the primary motivations behind this transition from human divers to ROVs
% The two primary drivers of this shift away from using human divers are safety and cost~\cite{mignotte7666smart}.
% Inspection of aquatic environments is challenging, especially in GPS-denied and cluttered underwater surroundings, such as pipeline inspection~\cite{zhao2022offshore}, monitoring of nuclear storage facilities~\cite{griffiths2016avexis} and liquid storage tank inspections~\cite{duecker2019learning}. 
% Precision movement is crucial in these environments as vehicles must navigate through narrow passages and avoid obstacles~\cite{lv2014design}.

\subsection{Accuracy requirements in constrained environments}
Navigating underwater vehicles through highly constrained physical environments is challenging and requires precise movement.
This is in contrast to operating in open oceans where robots generally move in free space and therefore the precision and accuracy requirements can be relaxed.
The accuracy and precision of a robot’s positioning and pose estimation impose a fundamental limitation on the performance of any navigational control system, since a robot cannot navigate with higher accuracy than its state estimation allows.

While the accuracy requirements of a positioning system will vary depending on the mission and environment,
it is useful to have quantitative targets, even if they can only be approximate.
To gauge the accuracy requirements for navigation, a representative example of a small underwater vehicle navigating through an opening that is 500~mm wide is considered.
Assuming that the robot is 340~mm wide~\cite{BluRoboticsbluerov2}, this would leave 80~mm either side to account for both position and control errors.
In a recent challenge statement from the UK nuclear industry~\cite{1234567}, an accuracy requirement of 50~mm was specified for revisiting the same position in a small, enclosed storage pond of 7~m x 7~m.
Therefore, it is concluded that positioning accuracy in the 0-100~mm range would be acceptable for many common missions in constrained underwater environments.
The accuracy is defined here as the euclidean distance between the estimated position and the actual position.
%
% However, existing positioning technologies for constrained underwater environments fail to achieve an optimal balance among environmental characteristics: system error, additional infrastructure, turbidity, range, salient features, and monetary cost, implying that there are always one or two significant shortcomings.

\subsection{Infrastructure and coverage}
To be useful in practical situations, a confined space positioning system should require minimal infrastructure and be capable of good coverage of the environment.
Fixing infrastructure, such as installing underwater cameras, beacons or markers, in the environment is time consuming, expensive and generally requires lengthy calibration.
%
% Moreover, in many situations, installing infrastructure is not feasible. For example, access might be highly restricted for safety reasons, in which case it would be impractical to install sensors or markers throughout the environment.
Moreover, installing infrastructure is often not feasible in environments where access is highly restricted due to safety concerns.
In such scenarios, deploying infrastructure becomes impractical.
A pertinent example is in nuclear fuel ponds, where safety protocols severely limit the introduction of external equipment.
% This highlights the challenges faced when attempting to utilize certain technologies in environments with stringent safety requirements.
It is also important that the system can operate over a high proportion of the environment, not suffer from blind spots, nor be confined to a local area.
Systems that rely on fixed infrastructure often suffer from such problems because the fixed equipment has restricted field of view and range.

%%%%%%%%%%%%%%%%%% CHAPTER 2 %%%%%%%%%%%%%%%%%%
\section{Review of Underwater Localisation Systems}\label{sec:lr}
The reason that positioning robots underwater remains particularly challenging relates mostly to the properties of water itself.
Technologies that are commonly used in air, such as GPS and LiDAR, rely on electromagnetic frequency bands that are highly attenuated by water, rendering them largely unusable.
There are some exceptions to this.
For example, there is a relative reduction in attenuation of visible light frequencies (380-750~nm) and as a consequence, in aquatic environments, cameras are the most successfully used sensor that relies on propagation of electromagnetic waves.
Aquatic applications that require information to be transferred over long distances typically use acoustic signals, which are not significantly attenuated in water.
%
% However, acoustic signals are subject to relatively slow propagation, low frequency, and are impacted by multipath issues~\cite{stojanovic2009underwater,singer2009signal}, these factors limit their accuracy and refresh rate when used for underwater positioning.
%
However, in highly constrained environments, multipath problems are exacerbated, rendering many acoustic positioning systems unsuitable~\cite{horri2020underwater}, particularly those that operate in lower frequency bands within which signals suffer less attenuation and therefore echoes dissipate slowly.

\subsection{Acoustic positioning}
The most widely used underwater positioning technology is based on acoustic triangulation and there are several standard system configurations available for its use in marine field robotics.
The main difference between the different setups is the distance between acoustic transponders (termed the baseline) and whether transponders are mounted to the seabed or to a surface ship.
For use in highly constrained environments, ultra short baseline (USBL) systems would be the most appropriate as they are designed for lower ranges and do not require transponders to be fixed to the infrastructure. However, typical accuracy of USBL systems is relatively low, 3-5\% of the range~\cite{sonardyneusbl}, which equates to up to 0.5~m over 10~m, and so is insufficient for the previously mentioned applications, with the additional problem that refresh rates will be relatively slow.

Sonar based simultaneous localisation and mapping (SLAM) is another acoustic technique that is widely reported in the literature~\cite{mcconnell2022overhead,ling2023active,teixeira2019dense}.
%
% Sonar SLAM in an underwater environment is analogous to LiDAR based SLAM in a terrestrial environment~\cite{oliveira2021feature}.
%
The technique has been reported to achieve positioning errors of 0.2~m over a 2.5~km trajectory, when used in combination with both IMU and DVL sensors~\cite{ozog2016long}.
Although this is an improvement, it is still insufficient.
More importantly, the quoted figures are from open water studies and the systems are unproven in confined industrial aquatic environments.

% The main limitation with sonar SLAM is the significant noise interference. Multipath echoes, resulting from reflections in constrained underwater environments, can create "phantoms" on sonar images, greatly increasing the difficulty of image processing and in extracting accurate positional information. A further consideration is that Sonar SLAM systems rely on costly measurements. 

% Aside from accuracy, a key issue that hinders the use of Sonar SLAM  that the update rate is slow and the measurements are noisy compared to LiDAR.
% %
% This means that maps take time to build and vehicles must move slowly to stop the robot loosing its place in the map.

\subsection{Vision based positioning}
 
The most accurate underwater positioning systems use several cameras, fixed to the perimeter of an environment, to track an array of markers that are fixed to the robot.
The commercial underwater motion capture system produced by Qualisys~\cite{qualisysycam}, for example, achieves sub-centimeter accuracy, low latency and fast refresh rates of 100~Hz.
Despite this impressive performance, such systems are typically more suited to lab settings, as they have significant setup, calibration and infrastructure requirements, are highly sensitive to water clarity, as well as having limited volume coverage.
Duecker~et~al.~\cite{duecker2020towards} inverted this principle, using a single camera and many marker objects.
They placed an array of 63 artificial markers around the perimeter of a tank and used a vehicle mounted camera, combined with AprilTag tracking, to estimate the pose of the camera, which is fixed on an underwater robot.
Although cheaper and able to cover a greater proportion of the environment than a system with wall mounted cameras, placing and maintaining many markers at known locations in the environment is not a practicable solution.

Underwater vision-based SLAM using onboard cameras is also a common solution, with at least one commercial product~\cite{vaarstsubslamx2} and several examples in the literature~\cite{joshi2019experimental}.
Vision-based SLAM depends heavily on recognizing and tracking salient environmental features and reliance on such features is a problem that has been reported when using aerial vehicles in GPS denied environments~\cite{chowdhary2013gps}.
As discussed above, vision penetration is reduced underwater, particularly in turbid waters, and causes image sharpness and visibility range to be reduced, which introduces significant challenges when using vision-based SLAM in aquatic applications.
Studies have confirmed that the performance of vision-based SLAM used underwater is inferior to that in air due to the low contrast of underwater images. In most cases, features are difficult to extract and are highly dependent on environmental conditions~\cite{zhao2020detecting}.

\subsection{Summary and Identified Research Gaps}
\textcolor{blue}{A review of recent advances in underwater localisation and navigation has been provided in this section. Table I summarises and compares the underwater localisation technologies reviewed in the previous subsections, which shows a simplified feasibility classification for each of the technologies in a confined underwater environment. The comparison focused on 6 main factors that evaluated if it can be deployed in confined spaces, including positioning accuracy, infrastructure requirement, turbidity, operating range, the necessity for a featurerich environment, and cost.}

\textcolor{blue}{Vision-based positioning systems generally offer high accuracy; however, their performance is often constrained by short sensing ranges and susceptibility to turbidity. These limitations can be partially mitigated through sensor fusion techniques that incorporate complementary modalities. The resulting positioning accuracy is influenced by both the characteristics of the integrated sensors and the specific fusion algorithms employed. For example, combining stereo vision, imaging sonar, and inertial measurement units (IMUs) can enhance overall system robustness. Nevertheless, in underwater scenarios with poor visibility or sparse environmental features, the accuracy largely depends on the performance of sensors such as imaging sonars or IMUs. Under these conditions, the inherent limitations of sonar-based methods, including degraded performance in cluttered or reflective environments, become more pronounced.}

\textcolor{blue}{Some underwater localisation systems are capable of achieving high accuracy without accumulating drift over time. However, these typically rely on external infrastructure. For instance, commercial optical motion capture systems like Qualisys require pre-deployment of underwater cameras, which may be impractical or infeasible in many confined or dynamic underwater environments. Similarly, electromagnetic localisation techniques used in LBL, SBL, and USBL systems, as well as methods dependent on fiducial markers or other fixed landmarks, face comparable deployment challenges.}

\textcolor{blue}{Thus, although existing technologies may perform well in isolated aspects of localisation, such as accuracy or range, there is no well-balanced solution that meets all the requirements for confined underwater environments.}
\subsection{Contribution}
In this work, a first-of-a-kind collaborative aquatic positioning (CAP) system, which aims to satisfy the requirements defined earlier, is proposed and evaluated experimentally in a typical industrial liquid storage tank (see MOVIE 1~\footnote{Please refer to the MOVIE-1 Introduction in the uploaded Supplementary Material.}).
The fundamental concept behind the CAP system is inspired by the mother-ship model, used in open oceans, where a surface vehicle with it's own sensor suite is used to help localise a subsurface vehicle in a global coordinate frame.
The key difference being that, in this work, the collaborating surface vehicle is highly mobile and able to move autonomously, staying above the subsurface vehicle.
%
% The surface vehicle can self-localise relative to features above the water surface and autonomously follow the underwater vehicle.
%
By combining information from sensors that are fixed to both the surface and underwater vehicles, and tracking a fiducial marker onboard the underwater vehicle, the position of the underwater vehicle can be determined.

Using a collaborative autonomous surface robot in this way has several benefits.
First, the majority of the translation from the origin coordinate system to the underwater vehicle frame is performed in air using an accurate LiDAR based approach;
only the direct translation down to the underwater vehicle is performed in water.
This means that camera based underwater localisation, which is the fastest and most accurate underwater technique, is appropriate due to the relatively short distances involved.
Second, coverage of the environment is almost complete because the camera follows the underwater vehicle and actively keeps it in the field of view.
Third, the system does not require any fixed infrastructure, and calibration is as simple as choosing the location of the reference coordinate system origin on the water surface.

Two variants of the CAP system are proposed in this paper: CAP\nobreakdash-CPnP, which uses camera based object tracking and a Perspective-n-Point (PnP) algorithm; and CAP\nobreakdash-CD, which uses a novel formulation to combine camera based object tracking and a pressure sensor on the underwater vehicle.
Both variants are evaluated experimentally during an autonomous underwater mission, with position data from the CAP system being produced in real-time and fed back to the underwater vehicle (in real-time) to enable an autonomous (waypoint guided) mission.

\textcolor{blue}{
While real-world underwater environments may involve visual occlusion due to clutter, equipment, or structural elements, this study focuses specifically on evaluating the core positioning performance of the proposed system under ideal line-of-sight conditions. The challenge of occlusion mitigation, including strategies such as multi-vehicle collaboration, sensor redundancy, or onboard fallback localisation, is considered an important direction for future extensions.
}

\begin{table*}[ht]
    \centering
    \caption{Summary of the most common underwater localisation technologies. Each positioning method's comparative results are derived from the most representative research achievements or commercial products using that method. The one with the highest positioning accuracy in this category is highlighted in bold.(The references for the 'Technology/Inventor' column are, in top-to-bottom order, as follow:~\cite{s2c})~\cite{kongsberg}~\cite{waterlinked}~\cite{topini2020lstm}~\cite{mur-artal2017orb}~\cite{westman2019degeneracy}~\cite{rahman2022svin2}~\cite{qualisys}~\cite{xing2021multi}~\cite{duecker2019integrated}}
    \includegraphics[width=\textwidth]{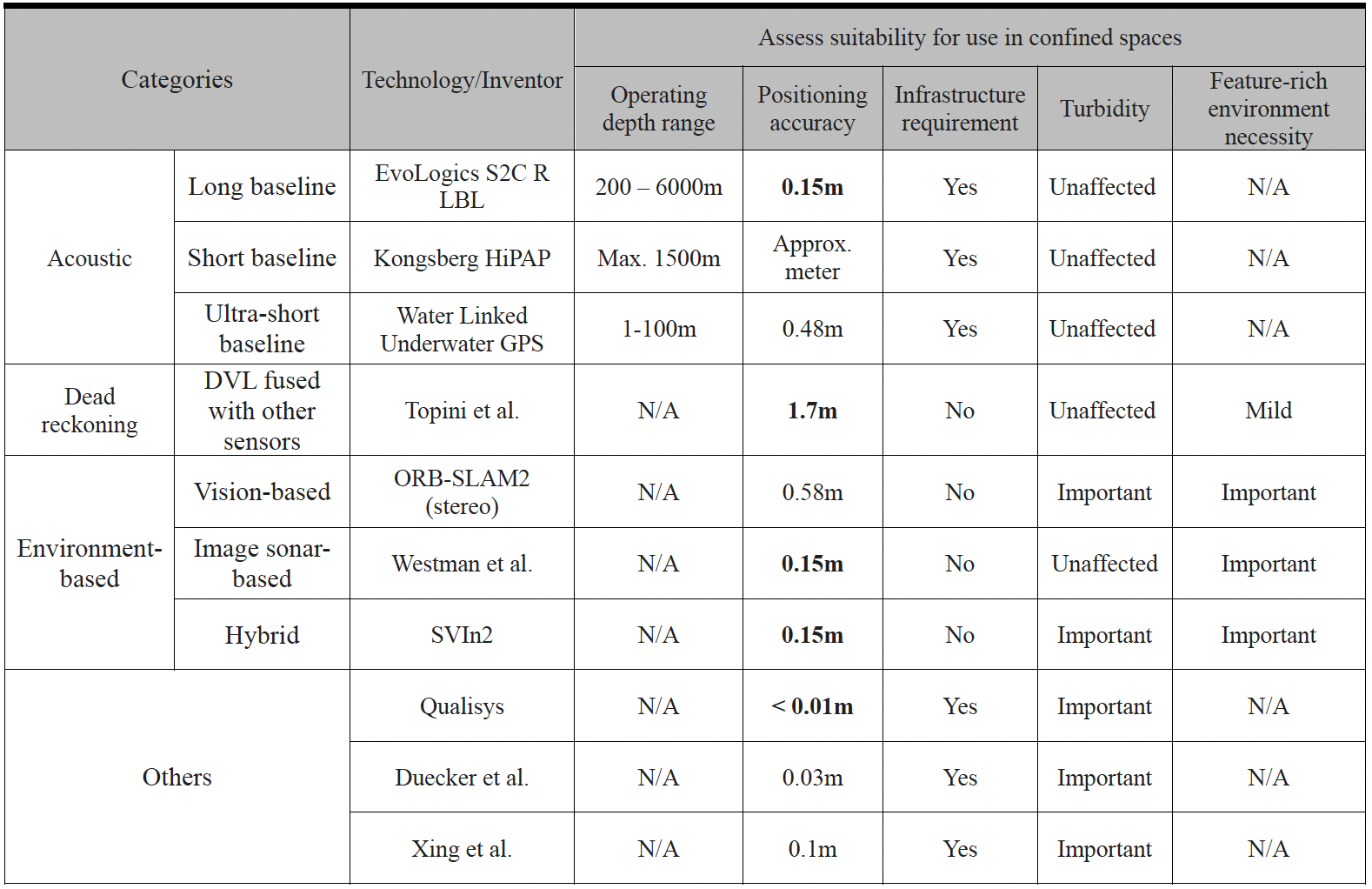}
    \label{tab:table_ramv2}
\end{table*}
% \begin{figure}[ht]
%     \centering
%     \includegraphics[width=\columnwidth]{figures/movie.png}
%     \caption{Movie 1.Summary of CAP system}
%     \label{fig:movie}
% \end{figure}

%%%%%%%%%%%%%%%%%% CHAPTER 3 %%%%%%%%%%%%%%%%%%
\section{Materials and Methods}\label{sec:mm} 
\subsection{Overview}
% Paragraph explaining the system, robots, sensors etc.

%
The two variants of the CAP system, namely \mbox{CAP-CPnP} and CAP\nobreakdash-CD, share several components such as the estimation of the surface robot's 6 degree of freedom (DOF) pose and use of the AprilTag to identify camera pixels that represent the corners of a fiducial marker.
The key difference between the two systems is that CAP\nobreakdash-CD does not require multiple pixels to be identified at known locations to enable use of a PnP algorithm~\cite{lepetit2009ep}.
Instead, only a single pixel needs to be identified, which broadens the horizon of image processing techniques that can be applied.
However, removing the PnP solver means that there is no longer a serial transform chain.
Therefore additional sensing as well as a new mathematical formulation are required to enable full and direct calculation of the underwater robot's pose.

\subsection{Hardware architecture - robotic platforms and sensors}

% \begin{figure}[htbp]
% \centering
% \includegraphics[width=\columnwidth]{figures/MallARD-JFR.pdf}
% \caption{\textbf{Hardware and system architecture.} (\textbf{A}) Components (\textbf{B}) MallARD 003 platform dimensions (\textbf{C}) BlueROV2 equipped with a depth/pressure sensor}
% \label{fig:MallARD}
% \end{figure}
% \begin{figure}[htbp]
% \centering
% \includegraphics[width=\columnwidth]{figures/MallARD.png}
% \caption{\textbf{ Hardware and system    architecture specifics.} (\textbf{A}) Components description. (\textbf{B}) MallARD\_003 electronic architecture.  (\textbf{C})}
% \label{fig:MallARD}
% \end{figure}
The hardware used in this study represents one possible physical incarnation of the positioning systems. As would be expected, the underlying mathematical formulations are agnostic to the choice of sensing methods and robots and there are several possible configurations. However, detail has been given below to facilitate understanding of the systems.

\begin{figure*}[ht]
    \centering
    \includegraphics[width=0.8\textwidth]{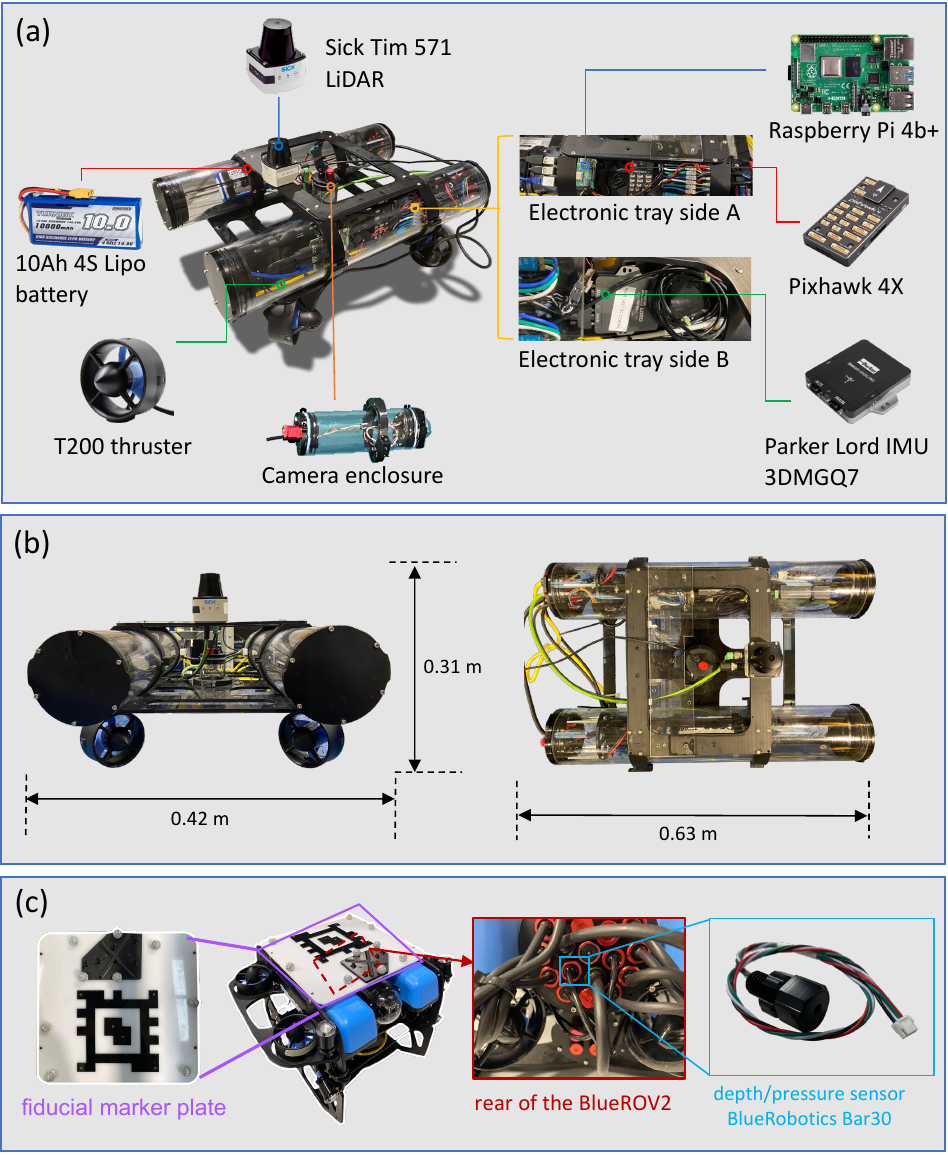}
    \caption{Hardware and system architecture.
    \textbf{(a)}: MallARD's salient components and their layout.
    \textbf{(b)}: MallARD's platform dimensions
    \textbf{(c)}: Customised BlueROV2 equipped with a depth/pressure sensor}
    \label{fig:mallard-split1}
\end{figure*}

The aquatic surface robot used in the proposed positioning system is MallARD (sMall Autonomous Robotic Duck) platform~\cite{groves2019mallard}, which is shown in Figure~\ref{fig:mallard-split1}.
% \begin{figure*}[ht]
%     \centering
%     \includegraphics[width=\textwidth]{figures/MallARD_split1-TFR.pdf}
%     \caption{Hardware and system architecture.
%     % 
%     \textbf{A}: MallARD components and their layout.
%     % 
%     \textbf{B}: MallARD 003 platform dimensions}
%     \label{fig:mallard-split1}
% \end{figure*}
% 
The dual pontoon configuration of MallARD ensures stability and also creates space at the robot's centre for sensor payloads.
To facilitate locomotion, MallARD is equipped with four bidirectional Blue Robotics T200 thrusters.
The thrusters are in a 45-degree configuration relative to the \textit{x} or \textit{y} axis, which allows vectoring in the robot's \textit{x} and \textit{y} axes and rotation about the robot's \textit{z}-axis.
MallARD has an on-board computer and is Robotic Operating System (ROS) enabled.
Motion commands are sent from ROS over a serial connection to a control unit (Pixhawk), which generates pulse-width modulation (PWM) signals that are sent to the electronic speed controllers (ESCs).
The ESCs in turn provide a phased output to the four brushless motors that drive the thrusters.

For use in the CAP system, MallARD was modified to include a downward facing low light HD camera and an inertial measurement unit (IMU).
\begin{figure}[ht]
    \centering
    \includegraphics[width=\columnwidth]{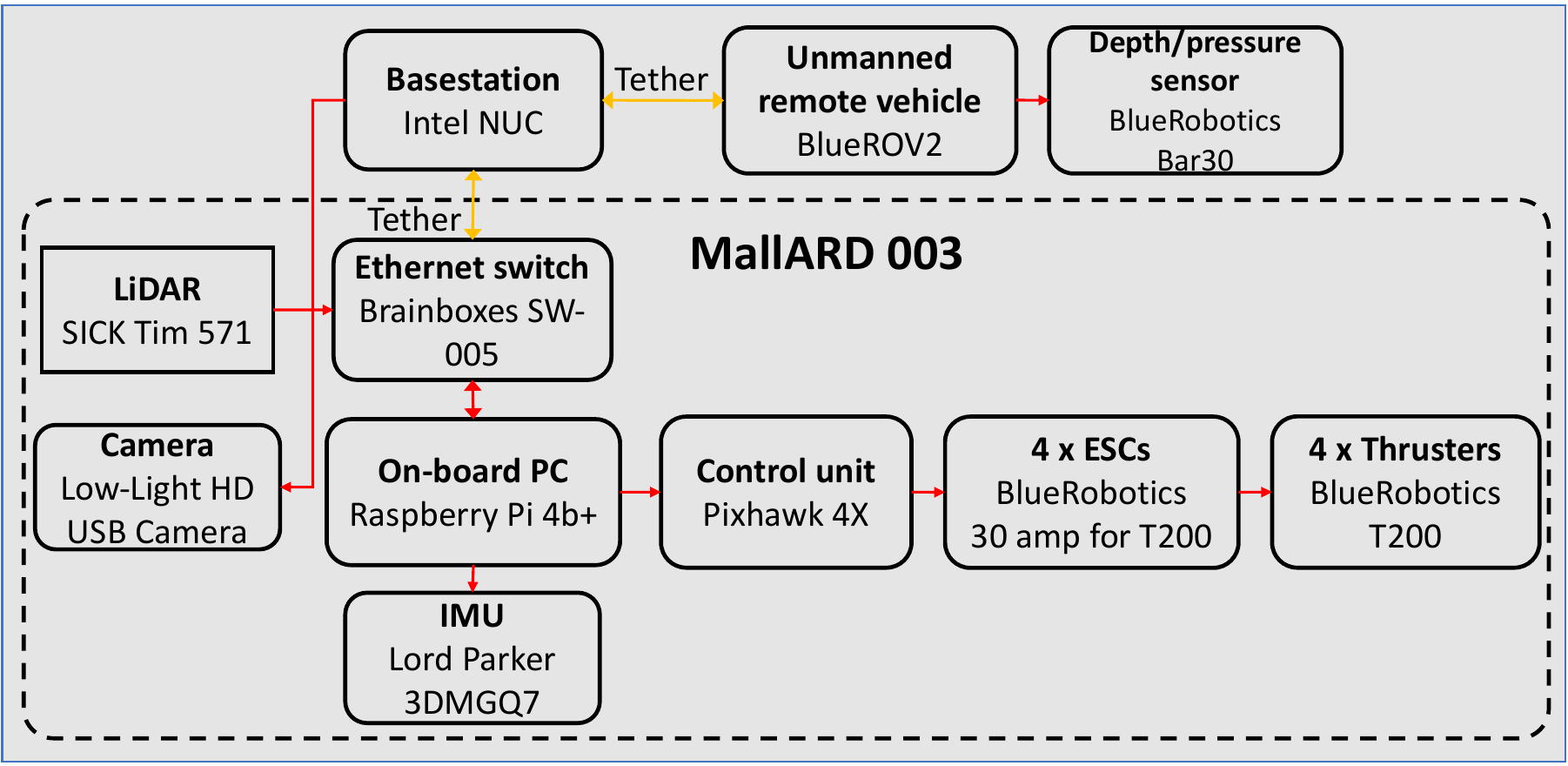}
    \caption{Schematic block diagram showing MallARD's electrical architecture}
    \label{fig:e-architecture}
\end{figure}
The modified robot layout and dimensions are depicted in Figure~\ref{fig:mallard-split1} (a) and (b) respectively, while Figure~\ref{fig:e-architecture} shows the electrical connections of the full system.
The camera was included to provide the surface robot with a clear video stream directly beneath the robot to enable tracking of the underwater robot.
This camera is a low-light HD~USB Camera, which was mounted in a waterproof enclosure in the robot's central payload area.
The IMU was added to enable full pose estimation of MallARD relative to the fixed environment.
%

 %
% MallARD is a holonomic autonomous surface vehicle composed of seven main parts: (I) chassis plates, (II) pontoons, (III) electronics tray, (IV) battery tray, (V)camera enclosure (VI) four thrusters, and (VII) waterproof 2D LiDAR. 
%

% MallARD is powered by a 14.8V, 10Ah Lithium polymer battery, housed in the left pontoon.
%

% , path planning, control, and underwater visual positioning. All computational components are housed in the right pontoon, All computational components are housed in the right pontoon, which includes an Park Lord 3DMGQ7 IMU connected to the on-board PC (low-cost RaspberryPi 4b+), utilized for more accurate measurement of the surface robot's pose.. 

The underwater robot used in this study is a commercially available BlueROV2, which has been customized to operate using ROS.
A fiducial marker, constructed using laser-cut acrylic sheet material, is fixed on top of the BlueROV2, as depicted in Figure~\ref{fig:mallard-split1}(c).
This marker is used to track the underwater robot in the field of view of the downward facing camera fixed to MallARD.
%
% It is important to note that 
%
\textcolor{blue}{The depth information was obtained using a BlueRobotics Bar30 pressure sensor, which was rigidly mounted at the rear of the underwater robot. The sensor was directly integrated into the ROS framework and provided depth measurements at a frequency of 15~Hz. Prior to deployment, the sensor was calibrated to ensure measurement accuracy (Appendix~\ref{appendix:depth_cali}).}

% Additionally, a pressure sensor, positioned at the BlueROV2's rear and illustrated in Figure~\ref{fig:mallard-split1}(c), has been integrated. This calibrated sensor provides accurate depth measurements (See Supplementary material~\ref{appdex:camaera_cali}).
% \begin{figure}[h]
%     \centering
%     \includegraphics[width=\columnwidth]{figures/MallARD_split2-TFR.pdf}
%     \caption{Customised BlueROV2 equipped with a depth/pressure sensor}
%     \label{fig:mallard-split2}
% \end{figure}

\subsection{\textcolor{green}{Homogeneous transforms}}
Given $\mathbf{p}^1_X$ that represents a point labelled $X$ $\in \mathbb{R}^3$ in coordinate frame $\mathcal{F}_1$, 
the coordinates of the same point can be represented in a different coordinate frame $\mathcal{F}_0$, given the transform from $\mathcal{F}_0$ to $\mathcal{F}_1$.
This coordinate frame transform can be expressed using a 4$\times$4 homogeneous transform matrix $\mathbf{H}^0_1$, which represents the pose of $\mathcal{F}_1$ with respect to $\mathcal{F}_0$
\begin{equation}
    \begin{bmatrix}
        \mathbf{p}^0_X \\
        1
    \end{bmatrix}
    = \mathbf{H}^0_1
    \begin{bmatrix}
        \mathbf{p}^1_X \\
        1
    \end{bmatrix}\text{,}
\label{eq: point transform}
\end{equation}
where
\begin{equation}
\mathbf{H}^0_1=\left[\begin{array}{cc}
\mathbf{R}^0_1 & \mathbf{p}^0_1 \\
0 & 1
\end{array}\right]
\in SE(3)\text{,}
\label{eq: H breakdown}
\end{equation}
and $\mathbf{p}^0_{1} \in \mathbb{R}^3$ is the translation from the origin of $\mathcal{F}_0$ to the origin of $\mathcal{F}_1$, $\mathbf{R}^0_1 \in \mathrm{SO}(3)$ is the rotation matrix from $\mathcal{F}_0$ to $\mathcal{F}_1$.
Homogeneous transforms can be formed into serial chains. For instance, if there is a third coordinate frame $\mathcal{F}_2$ and the transform from $\mathcal{F}_1$ to $\mathcal{F}_2$ is given by $\mathbf{H}^1_2$, the transform from $\mathcal{F}_0$ to $\mathcal{F}_2$ can be obtained by right multiplication of the transform chain, in order from start frame to end frame:

\begin{equation}
\mathbf{H}^0_2 = \mathbf{H}^0_1 \mathbf{H}^1_2.
\label{eq: H serial chain}
\end{equation}

% This allows the coordinates of the point $\mathbf{p}$ in $\mathcal{F}_2$ to be expressed as:

% \begin{equation}
% \begin{bmatrix}
%     \mathbf{p}^2 \\
%     1
% \end{bmatrix}
% = \mathbf{H}^0_2
% \begin{bmatrix}
%     \mathbf{p}^0 \\
%     1
% \end{bmatrix}.
% \end{equation}

% $\mathbf{H} \in SE(3)$ represents the combination of rotation and translation in three dimensions, which can also be expressed as the pair $[\mathbf{R}, \mathbf{p}]$. 

\subsection{Coordinate frames}
% \begin{figure}[htbp]
% \centering
% \includegraphics[width=\columnwidth]{figures/frames_sr.png}
% \caption{\textbf{Schematic diagram of the principles of each part of the CAP system.} (\textbf{A}) Overview of the proposed CAP system. (\textbf{B}) The coordinate frames abstracted from the CAP system, along with the Plücker line employed in the \mbox{CAP-CD} formulation. (\textbf{C}) Simulated planar self localisation of the surface robot based on SLAM. C(i) is a simulated environment and C(ii) shows the laser scan, resulting body frame pose estimate and a map (the corner of a tank). (\textbf{D}) Tilting of the surface on the water surface due to the waves. D(iii) is a flow diagram illustrating use of Extended Kalman Filter used to determine roll and pitch from IMU measurements. (\textbf{E}) Optical tracking system utilising fiducial markers and camera.}
% \label{fig:system}
% \end{figure}

\begin{figure*}[ht]
    \centering
    \includegraphics[width=16cm]{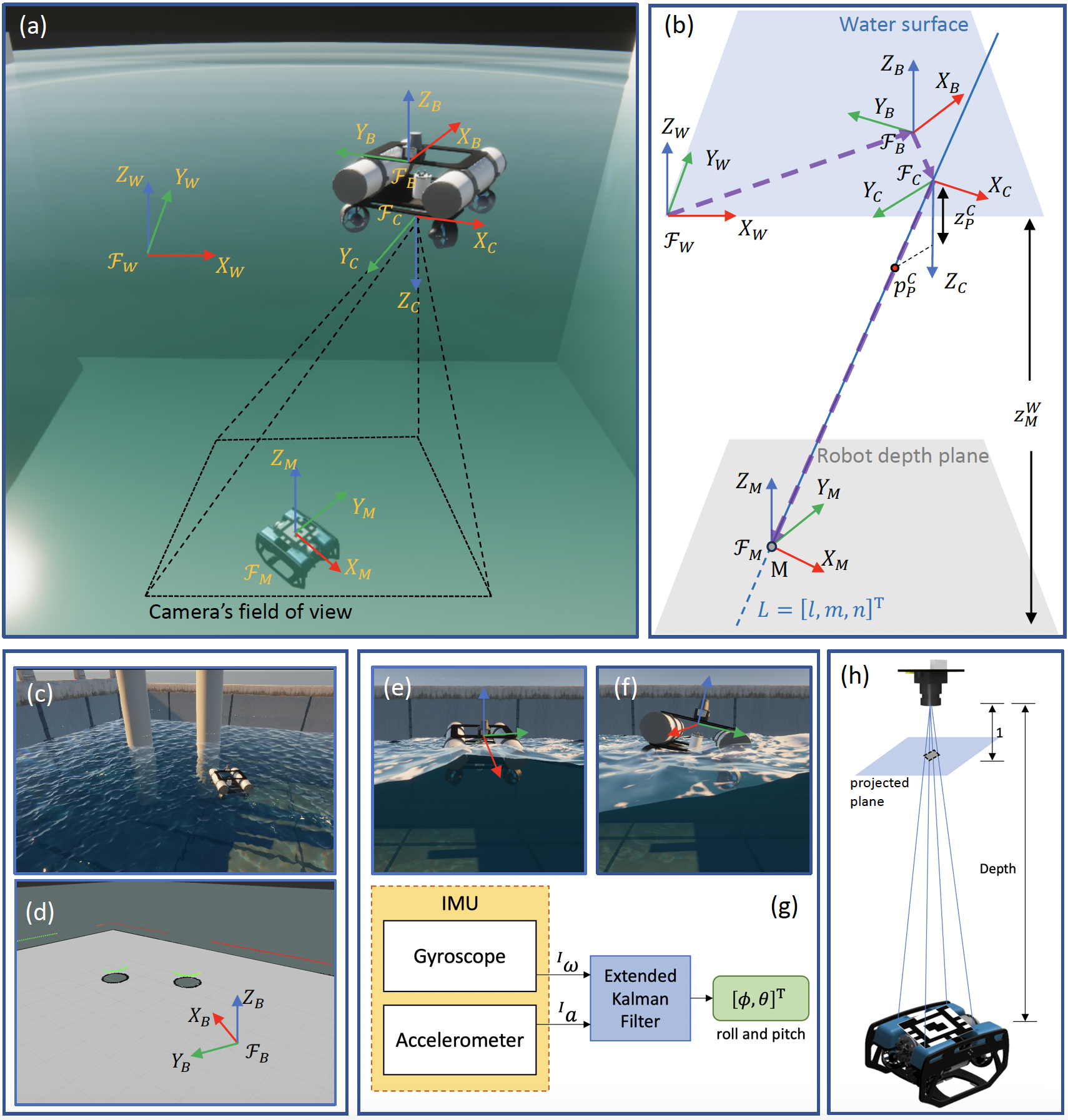}
    \caption{\textbf{Schematic diagram of the principles of each part of the CAP system.} \textbf{(a)} Overview of the proposed CAP system. \textbf{(b)} The coordinate frames abstracted from the CAP system, along with the Plücker line employed in the \mbox{CAP-CD} formulation. \textbf{(c)} is a simulated environment and \textbf{(d)} shows the laser scan, resulting body frame pose estimate and a map (the corner of a tank). \textbf{(e)} and \textbf{(f)}: Tilting of the surface on the water surface due to the waves. \textbf{(g)}: A flow diagram illustrating use of Extended Kalman Filter used to determine roll and pitch from IMU measurements. \textbf{(h)}: Optical tracking system utilising fiducial markers and camera.}

    \label{fig:math-1}
\end{figure*}

The coordinate systems involved in the design of each part of the CAP system are shown in Figure~\ref{fig:math-1} (a). The full positioning system is composed of the following coordinate frames: world frame~$\mathcal{F}_W$, MallARD baselink frame~$\mathcal{F}_B$, IMU frame~$\mathcal{F}_I$, camera frame (monocular)~$\mathcal{F}_C$ and marker frame~$\mathcal{F}_M$.
The world frame~($\mathcal{F}_W$) origin is assigned to a corner of the testing tank.
MallARD baselink~($\mathcal{F}_B$) is the geometric centre of MallARD.
The IMU frame~($\mathcal{F}_I$) is attached to the IMU and has a fixed transform from~$\mathcal{F}_B$.
The origin of the camera frame~($\mathcal{F}_C$) is located at the optical centre of the camera lens and also has a fixed transform from~$\mathcal{F}_B$.

\subsection{\mbox{CAP-CPnP} formulation} \label{sec: cap_pnp_formulation}

The CAP system aims to determine~$\mathbf{p}^W_M$: the position of the origin of the marker frame~$\mathcal{F}_M$ in the world frame~$\mathcal{F}_W$.
In the CAP-CPnP formulation, all elements of the serial transform chain can be determined independently; therefore, using Equations~(\ref{eq: point transform}) and~(\ref{eq: H serial chain}),~$\mathbf{p}^W_M$ can be calculated directly:
\begin{equation}
\left[\begin{array}{
c}
\mathbf{p}^W_M \\
1
\end{array}\right]=\mathbf{H}^W_B \mathbf{H}_C^B\left[\begin{array}{c}
\mathbf{p}^C_M \\
1
\end{array}\right].
\label{eq: CAP CPnP}
\end{equation}
The terms on the right hand side of Equation~(\ref{eq: CAP CPnP}) can be determined as follows.
From Equation~(\ref{eq: H breakdown}),~$\mathbf{H}^W_B$ is composed of~$\mathbf{R}^W_B$ and~$\mathbf{p}^W_{B}$.
$\mathbf{R}^W_B$ can be calculated using data from the IMU and 2D SLAM system on the surface vehicle, as detailed in Section~\ref{sec: MallARD's rotation}.
$\mathbf{p}^W_B$ can be constructed from the~$x$ and~$y$ components of the 2D SLAM output (Section~\ref{sec: 2D SLAM}), while the~$z$ component is assumed to be a static value that represents the offset between the world frame and MallARD's body frame.
Further detail regarding the method of obtaining $\mathbf{R}^W_B$ and $\mathbf{p}^W_{B}$ is given in the Section~\ref{sec:mm}-\ref{sec: 2D SLAM} and Appendix~\ref{appendix:tkf} respectively.
$\mathbf{H}_C^B$ is a measured static transformation from the ASV body frame~$\mathcal{F}_B$ to the camera frame~$\mathcal{F}_C$.
$\mathbf{p}^C_M$ is the position of the marker frame~($\mathcal{F}_M$) origin in the camera frame~($\mathcal{F}_C$) and is calculated using a fiducial marker tracking technique, which is a two step process.
First, the camera image is processed using AprilTag to detect the four pixel locations that relate to the four corners of a fiducial marker in the camera's image.
Second, the four pixel locations, together with the marker's dimensions and the camera's intrinsic matrix are used to determine the 6~DOF pose of the fiducial marker in the camera frame, from which~$\mathbf{p}^C_M$ can be extracted.
%]
This is a standard problem known as Perspective-n-Point (PnP) and has several available solutions, for example Direct Linear Transformation~\cite{hartley2003multiple} and Efficient PnP~\cite{lepetit2009ep}.
%
% Further detail regarding fiducial tracking and PnP usage is provided in the Supplementary Materials, Section~2.
% Those algorithm can return a pose (orientation and position) of AprilTag with respect to camera frame.

% that requires the camera's intrinsic matrix, the dimensions of the marker and the camera image

% given 3D points corresponding to 2D points on the image plane by first step， intrinsic matrix from a calibrated camera and dimension of the AprilTag to find the AprilTag's pose in camera frame is a classic topic called 

% 2D-3D correspondences is established by obtaining by considering the dimension of the tag and creating a set of 3D points corresponding to the corners of the tag. With the 2D-3D correspondences,  and camera intrinsic matrix, $\mathbf{p}^C_M$ can be calculated by functions for~\text {'solvePnP'}~\cite{opencvsolvepnp}.

Camera tracking systems have inherent noise due to their derivation from (necessarily) pixelated camera images.
Noise is exacerbated when 3-DOF translations are derived using Perspective-n-Point (PnP) methods.
This is primarily due to the low sensitivity of PnP solutions to depth variations ($z^C_M$).
Changes in depth cause relatively small changes to the image and, in turn, have a lesser effect on changing the pixels that are identified as corners of objects, causing low relative sensitivity and ambiguity.
Noise and ambiguity in $z^C_M$ mostly affects $z^W_M$, due to the fact that their associated axes are generally well aligned, however, noise in  $z^C_M$ also translates onto $x^W_M$ and $y^W_M$ when the ASV pitches and rolls.
To overcome the shortcomings of \mbox{CAP-CPnP}, CAP\nobreakdash-CD is proposed, which does not require the use of a PnP solver and, instead, incorporates a depth sensor onboard the underwater robot. More importantly, CAP\nobreakdash-CD does not require identification of multiple feature locations on the robot (tag corners in the case of fiducial markers), a single point in the camera image is sufficient.

\subsection{CAP\nobreakdash-CD formulation}
The CAP\nobreakdash-CD formulation does not use a PnP solver (or equivalent) and, therefore, $\mathbf{p}^C_M$ is undefined, breaking the transform chain.
However, with the inclusion of a pressure sensor on the underwater robot, that can be calibrated to measure water depth, it is possible to directly calculate~$\mathbf{p}^W_M$.
%

% \begin{figure}[h!]
%     \centering
%     \includegraphics[width=\textwidth]{figures/frames.png}
%     \caption{Abstract frames of the system. ($W$ - origin of world frame, $B$ - origin of body frame, $C$ - origin of camera frame, $M$ - origin of marker frame)}
%     \label{fig:frames}
% \end{figure}
Figure~\ref{fig:math-1} (b) gives a graphical representation of the method.
Consider a Plücker line that passes through the origins of~$\mathcal{F}_C$ and~$\mathcal{F}_M$, and a horizontal plane defined by the depth sensor measurement $z^W_M$.
By finding the intersection between the Plücker line and the horizontal plane, $\mathbf{p}^W_M$ can be calculated.

The Plücker line is defined by two points in the world frame~$\mathcal{F}_W$.
The first point $\mathbf{p}^W_C$ is the origin of~$\mathcal{F}_C$ which can be found using
\begin{equation}
\left[\begin{array}{c}
\mathbf{p}^W_C \\
1
\end{array}\right]=\mathbf{H}^W_B \left[\begin{array}{c}
\mathbf{p}^B_C \\
1
\end{array}\right]
\label{equ:2homos}.
\end{equation}

Since the origin of~$\mathcal{F}_M$ is unknown, another point on the line must be found for the line to be defined. 
To find this second point, AprilTag is used to identify the camera pixel locations that represent the corners of the  fiducial marker; these are averaged to give the single pixel location of the centre of the marker:~$u_p$ and~$v_p$.
The camera's intrinsic matrix is then used to identify the components $x^C_P$ and $y^C_P$ of a projected point~$\mathbf{p}^C_P$.
According to the definition of the intrinsic matrix, $z^C_P=1$ for all cases.
Therefore~$\mathbf{p}^C_P$ can be identified as follows:
\begin{equation}
\label{eq: xcp}
\mathbf{p}^C_P=
\left[\begin{array}{l}
x^C_P \\
y^C_P \\
1
\end{array}\right]=\left[\begin{array}{lll}
f_x & 0 & c_x \\
0 & f_y & c_y \\
0 & 0 & 1
\end{array}\right]^{-1}\left[\begin{array}{l}
u_{centre} \\
v_{centre} \\
1
\end{array}\right]\text{,}
\end{equation}
where the 3~$\times$~3 matrix is the camera's intrinsic matrix.
$\mathbf{p}^C_P$ lies on the the Plücker line that passes through the origin of~$\mathcal{F}_C$ and~$\mathcal{F}_M$.
However, for the Plücker line to be defined in~$\mathcal{F}_W$ the point must be transformed into~$\mathcal{F}_W$, the world frame:
\begin{equation}
\left[\begin{array}{c}
\mathbf{p}^W_P \\
1
\end{array}\right]=\mathbf{H}^W_B \mathbf{H}_C^B\left[\begin{array}{c}
\mathbf{p}^C_P \\
1
\end{array}\right].
\end{equation}

%%%%%%%%%%%%%%%%%%%%%%%%%%%%%%%%%%%%%%%%%%%%%%%%%%%%%%%%%%%%%%%%%%%%55
%%%%%%%%%%%%%%%%%%%%%%%%%%%%%%%%%%%%%%%%%%%%%%%%%%%%%%%%%%%%%%%%%%%%%5

% Imagine a line $\mathbf{\vec{L}}$ passing through both point C and point M.

Now, given $\mathbf{p}^W_C$ and $\mathbf{p}^W_P$ the Plücker line can be defined in the world frame.
In general, the equation of a line with direction vector $\mathbf{l}=[l, m, n]^{\top}$ that passes through the point $\left[x_1, y_1, z_1\right]^{\top}$ is given by the formula
\begin{equation}
\frac{x-x_1}{l}=\frac{y-y_1}{m}=\frac{z-z_1}{n}=k\text{,}
\end{equation}
% Both Point P, the tag's centre projected onto the scaled plane whose perpendicular distance to camera $x-y$ plane is 1, and C, the centre of camera lens, are on the same line.
where $k$ ranges over all real numbers and represents the position on the line.
By defining
\begin{equation}
\label{eq: lmn}
[l, m, n]^{\top}=\left[x_C^W-x_P^W, y_C^W-y_P^W, z_C^W-z_P^W\right]^{\top}
\end{equation}
the Plücker line can be expressed as:
% \begin{equation}
% \label{eq: plucker}
% \mathbf{r}=\mathbf{r}_{P^W}+\left(\mathbf{r}_{C^W}-\mathbf{r}_{P^W}\right) k
% \end{equation}
% where
% $\mathbf{r}$ is the position vector of any point on the line,
% $\mathbf{r}_{P^W}$ is the position vector of point~$\mathbf{p}^W$, $\mathbf{r}_{P^W}=\left(\begin{array}{c}x_P^W \\ y_P^W \\ z_P^W\end{array}\right)$,
% $\mathbf{r}_{C^w}$ is the position vector of point~$\mathbf{C}^W$, $\mathbf{r}_{C^w}=\left(\begin{array}{l}x_C^W \\ y_C^W \\ z_C^W\end{array}\right)$,
\begin{equation}
\label{eq: x(k)}
x=x_P^W+\left(x_C^W-x_P^W\right) k\text{,}
\end{equation}
\begin{equation}
\label{eq: y(k)}
y=y_P^W+\left(y_C^W-y_P^W\right) k\text{,}
\end{equation}
\begin{equation}
\label{equ:findk}
z=z_P^W+\left(z_C^W-z_P^W\right) k\text{.}
\end{equation}
Since $z_M^W$ can be found directly from the calibrated pressure sensor measurement, the value required for $k$ which effectively identifies the intersection between the horisontal plane given by $z_M^W$ and the Plücker can be computed by:
\begin{equation}
k=\frac{-z_M^W-z_P^W}{z_C^W-z_P^W}.
\end{equation}
Therefore, the tag's unknown coordinates $x_M^W$ and $y_M^W$ can be found by substituting $k$ back into Equation~(\ref{eq: x(k)}) and~(\ref{eq: y(k)}).

\subsection{Autonomous following}
%One paragraph about the following
For the CAP system to function, the marker on the underwater robot must be within the field of view of the  surface vehicle's downward facing camera.
To achieve this, a range of control techniques could be applied and in this work visual servoing was implemented~\cite{yao2023image}.

Initially, four target feature points that represent the desired position of the corners of the fiducial marker are defined on the projected image plane.
The fiducial marker tracking system then continuously detects these four points and compares them with the corresponding target projection points.
The aim of the visual servoing system is to minimize the difference between the desired and tracked positions.
This difference is translated into how the surface robot should move to ensure that the detected points match (or fall within an acceptable range of deviation) the target feature points, thereby enabling the surface robot to automatically follow the underwater robot.

%%%%%%%%
%%%%
%%%%%
% \begin{figure}[ht]
%     \centering
%     \includegraphics[width=\columnwidth]{figures/Figure-1_tfr2.pdf}
%     \caption{Simulated scenarios to aid description of self-localisation and tilting analysis of the surface robot:
%     % 
%     (a): MallARD in a simulated aquatic environment, and
%     % 
%     (b):The SLAM system visualisation including the multi coloured dots of the laser scan, the black lines of the map and a body frame pose estimate.
%     % 
%     (c) and (d): Tilting of the surface robot due to surface waves, and (e): Schematic diagram of  Extended Kalman Filter inputs and outputs.}
%     \label{fig:slam_imu}
% \end{figure}

\subsection{MallARD's 2D SLAM system}
\label{sec: 2D SLAM}

%self-positioning using LiDAR based on SLAM - $\boldsymbol{p^W_B}$ \& $\boldsymbol{\psi^W_B}$}
MallARD is equipped with a waterproof 2D LiDAR, enabling planar localisation relative to the pool walls.
Because the 2D LiDAR operates in a plane which is parallel to the water surface, the SLAM system provides positions $x^W_B$ and $y^W_B$ as well as yaw angle $\psi^W_B$ on that plane.
Because in this application there is no reliable source of odometry, the choice of 2D slam algorithms is limited.
In the current work, a customised version of Hector mapping is used as the SLAM framework as this does not require any odometry.
The customisations made allow the map to be locked, preventing corruption in longer trials; and allow the position and rotation to be output relative to a fixed reference location in the pool. By default, Hector mapping outputs data relative to the start location, which is impractical and non-repeatable for consistent testing.
Upon immersion and activation in the aquatic environment, MallARD transmits  LiDAR scans to the SLAM software.
To build a 2D map of the pool's walls, MallARD is manually navigated around the pool using a joystick.
Once the mapping phase reaches completion, the map is locked and MallARD's autonomous following mode is initiated. During this phase, the primary function of the SLAM algorithm is localisation, given that the spatial map undergoes no changes, as depicted in Figure~\ref{fig:math-1}(c) and (d).

% \begin{figure}[htbp]
%     \centering
%     \includegraphics[width=\columnwidth]{figures/surface_slam.png}
%     \caption{(a): MallARD in a simulated aquatic
% environment, and (b):The SLAM system visualisation including the multicoloured dots of the laser scan, the black lines of the map and a bodyframe pose estimate.}
%     \label{fig:surface_slam}
% \end{figure}

\textcolor{green}{Hector SLAM is currently used on the surface robot. Although there is no explicit analysis regarding its handling of dynamic obstacles, the probabilistic map cell update will update cells based on whether free or occupied space has been detected by the laser scan. By combining the pre-built map, the system can respond to dynamic changes in the environment.}

\subsection{MallARD's rotation relative to the world frame $\textbf{R}_B^W$}
\label{sec: MallARD's rotation}

As MallARD navigates through water, it undergoes roll and pitch due to resultant hydrodynamic forces and small waves on the water surface, as depicted in Figure~\ref{fig:math-1} (e) and (f).
For both CAP\nobreakdash-CD and \mbox{CAP-CPnP}, it is necessary to know \mbox{MallARD}'s body frame rotation  relative to the world coordinate system~$\mathbf{R}^W_B$.
For mathematical convenience $\mathbf{R}^W_B$ is calculated using Euler angles in the Z-Y-X sequence and then converted to a rotation matrix.
While the issue of gimbal lock is a known problem when using Euler angles, it is unlikely to occur in this case because rotations about the \textit{y}-axis generally remain within approximately 10 degrees of zero.

% \begin{figure}
%     \centering
%     \includegraphics[width=\columnwidth]{figures/tilting_and_ekf_block.png}
%     \caption{(a) and (b): Tilting of the surface robot due to
% surface waves, and (c): Schematic diagram of Extended Kalman Filter inputs and outputs.}
%     \label{fig:tilting_block}
% \end{figure}
%
% If the body frame rotation about the \textit{y}-axis comes near to 90 degrees (where gimbal lock occurs) there will likely be lager problems than gimbal lock.

To calculate~$\mathbf{R}^W_B$ in the Z-Y-X Euler angle form, the rotation about \textit{z}-axis  is decoupled from the rotations about \textit{y}-axis  and \textit{x}-axis .
Although \mbox{MallARD}'s IMU has a built in 3-axis compass, the compass-provided measurements are unreliable due to the magnetic fields generated by metallic structures and MallARD's own electronic equipment and motors.
Therefore, the magnetometer cannot provide a lock for the yaw measurement (rotation about \textit{z}-axis  in the Euler sequence).
However, yaw can be acquired through LiDAR-based SLAM and this is used as the \textit{z}-axis  component of the sequence.
This approach is valid because $z$ is the first rotation in the sequence and is therefore about the $z$-axis of~$\mathcal{F}_W$.
MallARD's roll and pitch (\textit{y}-axis  and \textit{x}-axis  rotations in the Euler sequence) must now be computed relative to the stabilised body frame, which is a version of the body frame without any roll or pitch.

The tilting extended Kalman filter (EKF) is used to find the second two rotations in the sequence.
As shown in Figure~\ref{fig:math-1}(g), the EKF takes two vector inputs, which are 3-axis angular rate $\boldsymbol{\omega} = [{\omega}_x, {\omega}_y, {\omega}_z]^{\top}$ measured by the gyroscope and 3-axis acceleration $\boldsymbol{a} = [{a}_x, {a}_y, {a}_z]^{\top}$ measured by the accelerometer, and outputs the rotations, which are pitch($\theta$) and roll($\phi$).
In this research, the motion of the USV does not exhibit prolonged substantial accelerations (other than gravitational acceleration) for an extended period of time.
Therefore, it is assumed that the acceleration vector is identical to the gravity vector.
Using the full Euler angle Z-Y-X sequence~$\mathbf{R}^W_B$ can be transformed to a rotation matrix using the following equation:

\begin{equation}
\mathbf{R_B^W}
=\left[\begin{array}{ccc}
c \psi c \theta & c \psi s \theta s \phi-s \psi c \phi & c \psi s \theta c \phi+s \psi s \phi \\
s \psi c \theta & s \psi s \theta s \phi+c \psi c \phi & s \psi s \theta c \phi-c \psi s \phi \\
-s \theta & c \theta s \phi & c \theta c \phi
\end{array}\right]
\end{equation}
%
% However, yaw can be acquired through LiDAR-based SLAM. In this case, Euler angles are calculated in IMU (pitch and roll) and SLAM (yaw) and then combined to determine \mbox{MallARD}'s rotation $\mathbf{R}^W_B$.

% Given that the tilting angles (pitch and roll) are obtained from the IMU and the yaw is sourced from SLAM, which are then integrated into a comprehensive three-dimensional rotation, the use of the Euler angle sequence of Z-Y-X is found to be convenient. For surface robots, when rotating in the Z-Y-X sequence, the rotation around the \textit{y}-axis will not reach 90 degrees, effectively preventing gimbal lock. 

%%%%%%%%%%%%%%%%%% CHAPTER 4 %%%%%%%%%%%%%%%%%%
\section{Rrsults}\label{sec:r} 
\subsection{Experiment setup}
\begin{figure}[ht]
\centering
\includegraphics[width=\columnwidth]{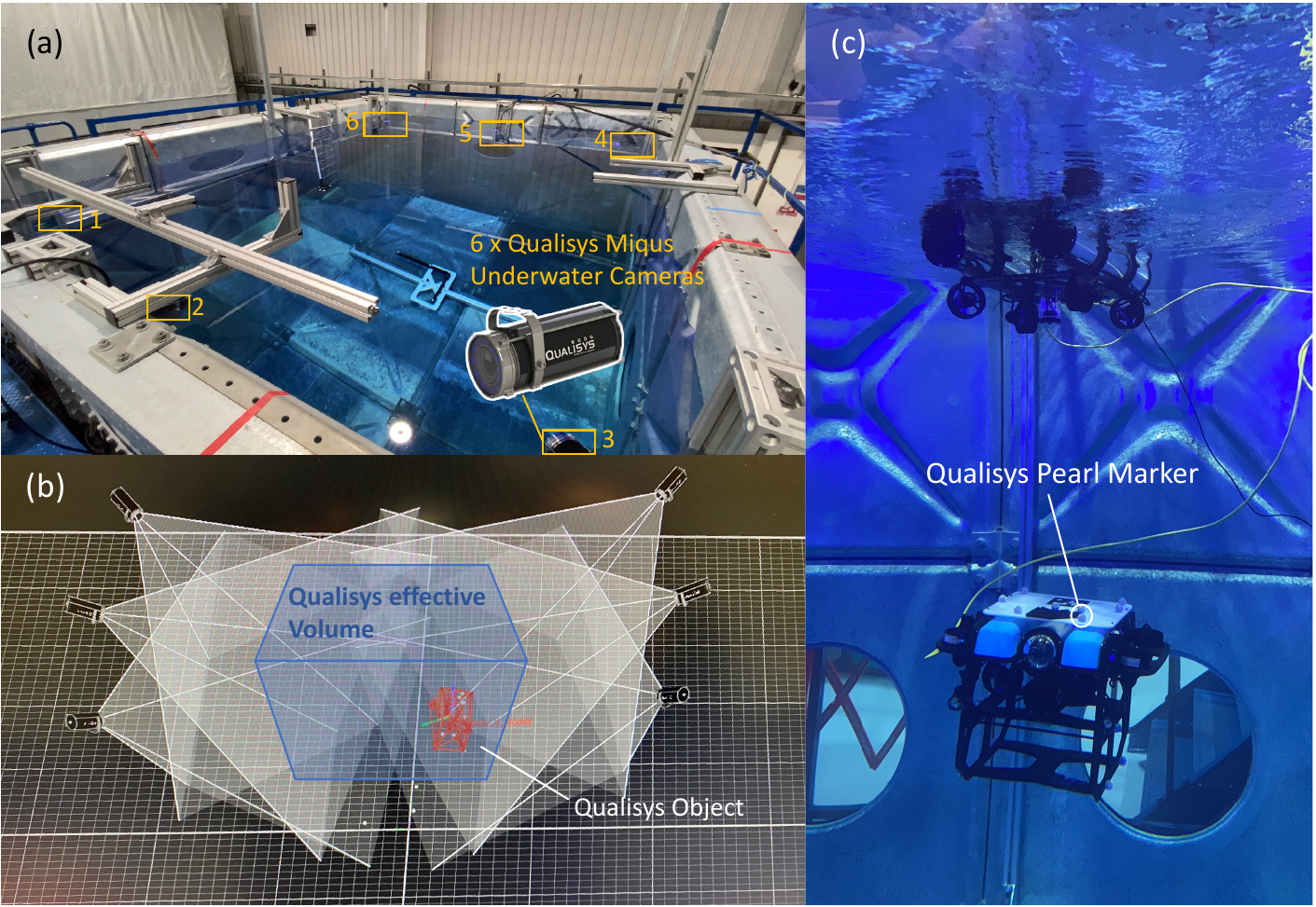}
\caption{Experimental field and setup. \textbf{(a)}: Overview of the experimental tank. \textbf{(b)}: Qualisys system setup and effective volume \textbf{(c)}: MallARD and BlueROV2 deployed in the experimental pond. BlueROV2 is mounted with pearl marker for Qualisys system tracking.}
\label{fig:Qualisys}
\end{figure}
The CAP system was evaluated using data collected in the 4.8$\times$3.6$\times$2.0$\,$m (length, width, depth) indoor test tank shown in Figure~\ref{fig:Qualisys} (a).
The positioning accuracy of the system was validated using a high-accuracy, 6 camera Qualisys Miqus M5 underwater motion tracking system.
The submerged Qualisys cameras, as shown in Figure~\ref{fig:Qualisys}(a), were fixed to the walls of the tank.
Due to field of view limitations, the Qualisys system could not cover the entire tank's volume, and as a consequence the experiments were conducted in a smaller region of the tank, as illustrated in Figure~\ref{fig:Qualisys} (b).
Qualisys tracking markers were placed on the BlueROV2 and on a customised marker plate that was used to allow the Qualisys object frame to be accurately located on to the BlueROV2.
Figure~\ref{fig:Qualisys} (c) shows the BlueROV2 with Qualisys pearl markers attached. When calibrated, the accuracy of the Qualisys system over the effective volume was 1~mm with regard to position and 0.1$^{\circ}$ for rotation.

% 3.6m $\times$ 4.8m $\times$ 2.4m 

% \subsection*{Data collection}
% In the experiments, the BlueROV2 was manually operated in the tank while maintaining a fixed pitch and roll. MallARD was also manually driven such that the fiducial marker on the BlueROV2 was kept in the field of view of MallARD's camera. MallARD will be programmed to follow the marker autonomously but the focus of the present research is to validate and quantify the accuracy of the proposed positioning system.

BlueROV2, MallARD, and the basestation all utilize ROS, enabling real-time data sharing and synchronization to a single clock, specifically the clock of the basestation.
All sensor data was generated and processed in real-time and recorded on the basestation.
Real-time pose data from the Qualisys system were also bridged into the ROS system and recorded on the basestation.

% In the CAP system, there are several estimated rotations and translations between different coordinate systems, such as between the camera frame and the USV's body frame ($\mathbf{R^C_B}, \mathbf{p^C_B}$), the Qualisys world frame and the world frame ($\mathbf{R^Q_W}, \mathbf{p^Q_W}$). To better analyze the sources of error in the CAP system, we employ Particle Swarm Optimization (PSO) to optimize these estimated parameters.

\subsection{SLAM on the Water Surface of the Experimental Tank}
\begin{figure}[htbp]
    \centering
    \includegraphics[width=\columnwidth]{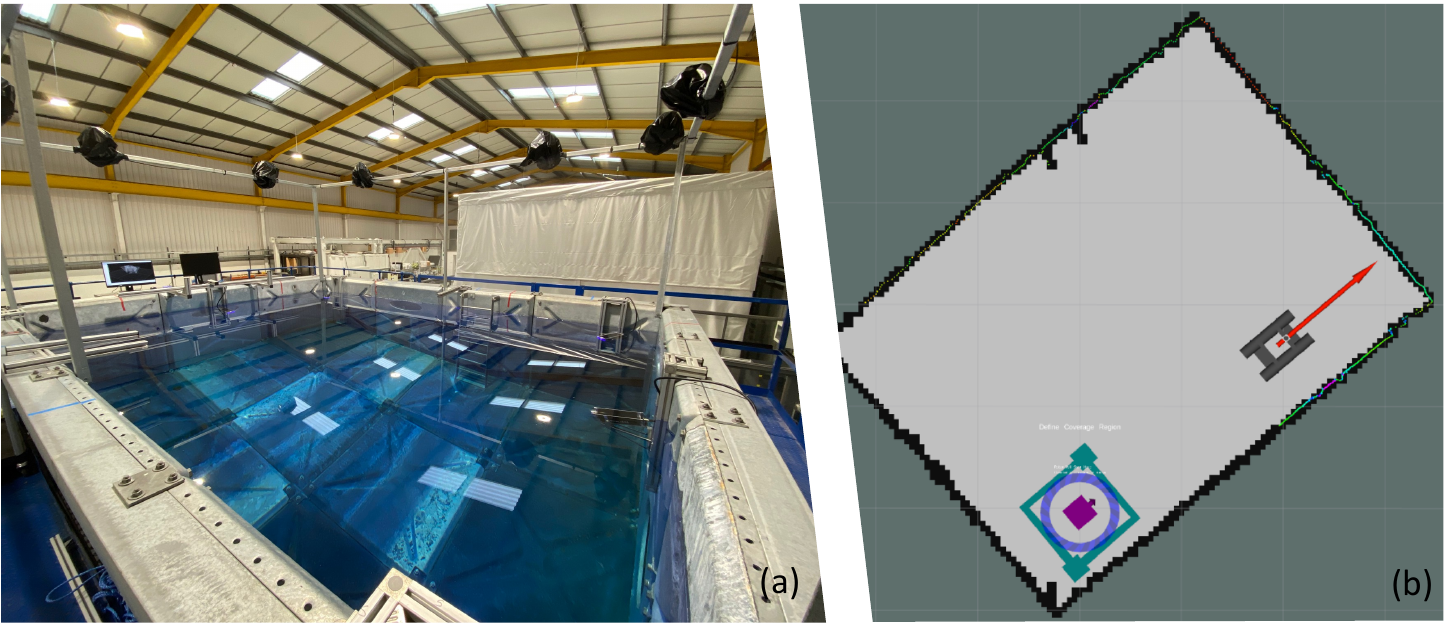}
    \caption{\textbf{(a)}:Mapping of the water surface area of the experimental tank. \textbf{(b)}: Using Hector SLAM, displayed in RViz.}
    \label{fig:raico_tank_map}
\end{figure}

As an essential component of the CAP system, the self-localisation of the surface robot requires SLAM to be performed on the area above the water surface of the experimental tank. In this work, Hector SLAM was employed as the localisation algorithm for the surface robot. After the map was completed, it was locked to prevent further updates. This ensured that the surface robot's localisation remained accurate, avoiding map failure and localisation errors during rapid or intense movements of the robot. The area above the water surface of the experimental tank, primarily the edges of the tank's walls, is shown in the Fig~\ref{fig:raico_tank_map}.

\subsection{Experimental validation in the test tank}

% \begin{figure*}[htbp]
%     \centering
%     \includegraphics[width=0.9\textwidth]{figures/result_tfr.pdf}
%         \caption{Overview of CAP system during testing.
%         % 
%         (a): Overlaid snapshots of CAP system; the trajectories of MallARD and BlueROV2 are shown as the green and red path.
%         % 
%         (b): 3D trajectory of BlueROV2 estimated by CAP-CD and \mbox{CAP-CPnP} against ground truth overtime respectively. BlueROV2 and MallARD operating in autonomous mode. The BlueROV2 is programmed to move in a specific trajectory. Meanwhile, the two robots of the CAP system operate in a leader-follower configuration (BlueROV2 as leader and MallARD as follower).}
%     \label{fig:result}
% \end{figure*}

\begin{sidewaysfigure*}[htbp]
    \centering
    \includegraphics[width=0.9\textwidth]{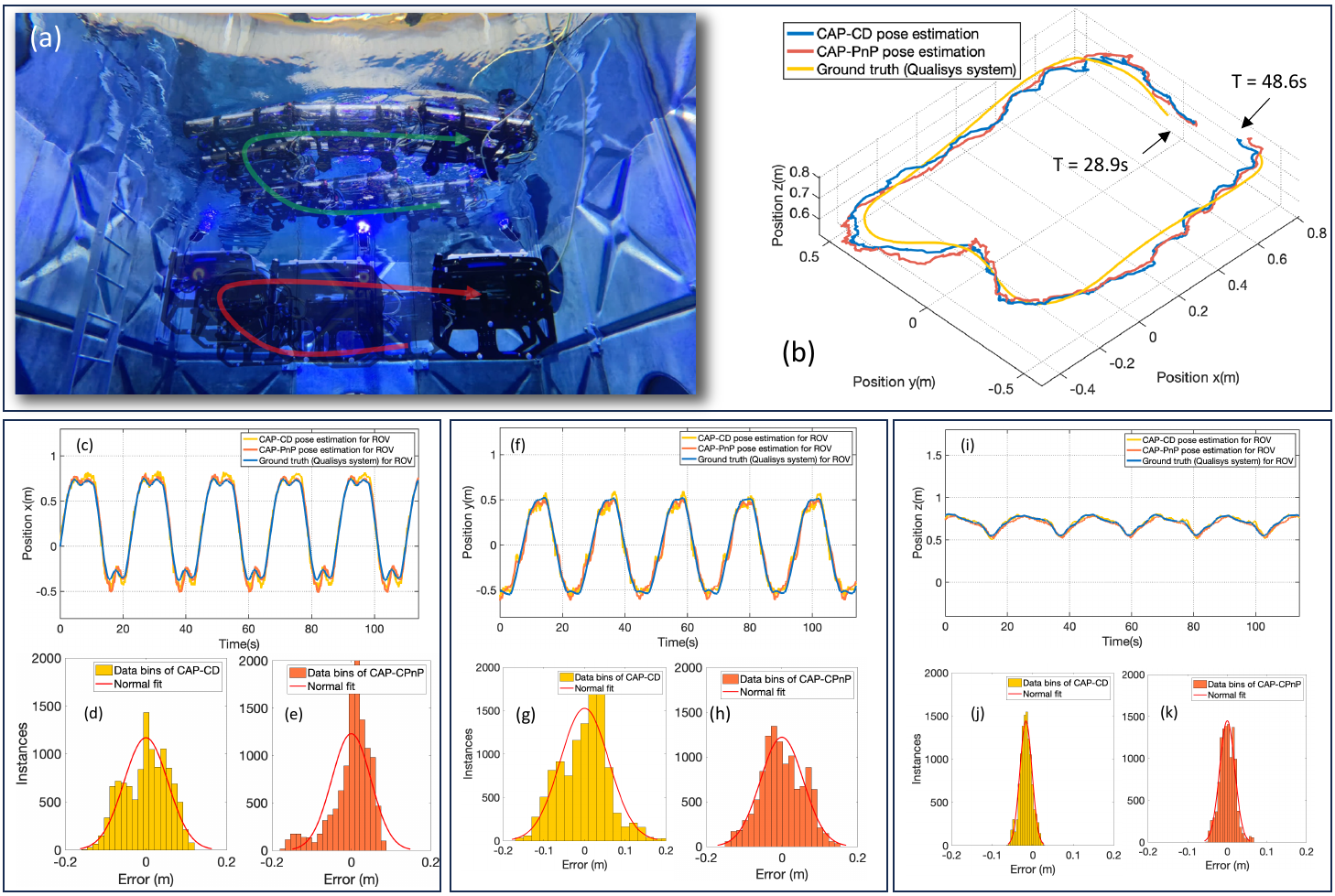}
        \caption{Overview of CAP system during testing.
        (a): Overlaid snapshots of CAP system; the trajectories of MallARD and BlueROV2 are shown as the green and red path.
        (b): 3D trajectory of BlueROV2 estimated by CAP-CD and \mbox{CAP-CPnP} against ground truth overtime respectively. BlueROV2 and MallARD operating in autonomous mode. The BlueROV2 is programmed to move in a specific trajectory. Meanwhile, the two robots of the CAP system operate in a leader-follower configuration (BlueROV2 as leader and MallARD as follower). 
        \textbf{(c)}, \textbf{(f)} and \textbf{(h)} The positioning of BlueROV2 by \mbox{CAP-CD} and \mbox{CAP-CPnP} in comparison with the ground truth along the X, Y, and Z axes, respectively. \textbf{(d)}, \textbf{(g)} and \textbf{(j)} The error histograms of \mbox{CAP-CD} on the X, Y, and Z axes. \textbf{(e)}, \textbf{(h)} and \textbf{(k)} The error histograms of \mbox{CAP-CPnP} on the X, Y, and Z axes.}
    \label{fig:result1}
\end{sidewaysfigure*}

The two positioning systems described in Section~\ref{sec:mm} were evaluated in this work, namely CAP-CD and CAP-CPnP.
To evaluate the performance of CAP system, the underwater robot was programmed to move autonomously along three pre\hbox{-}programmed trajectories.
These trajectories were: square, random and lawnmower pattern.
To comprehensively assess the impact of depth variations on the system, the depth of the underwater robot was varied by up to 1~m as it followed the pre\hbox{-}programmed trajectories.
Concurrently, the surface robot autonomously followed the underwater robot, ensuring that the fiducial marker, fixed on the underwater robot, remained within the field of view of the downward-facing camera on the surface robot. 
The validation experiment collected three sets of data for each of the square, random, and lawnmower patterns with relatively large depth variations (up to 1~m), with each dataset lasting for 120~seconds. Furthermore, the datasets for each of the square, random, and lawnmower patterns, were collected with minor depth fluctuations (up to approximately 0.3~m).

\begin{figure*}[htbp]
    \centering
    \includegraphics[width=\textwidth]{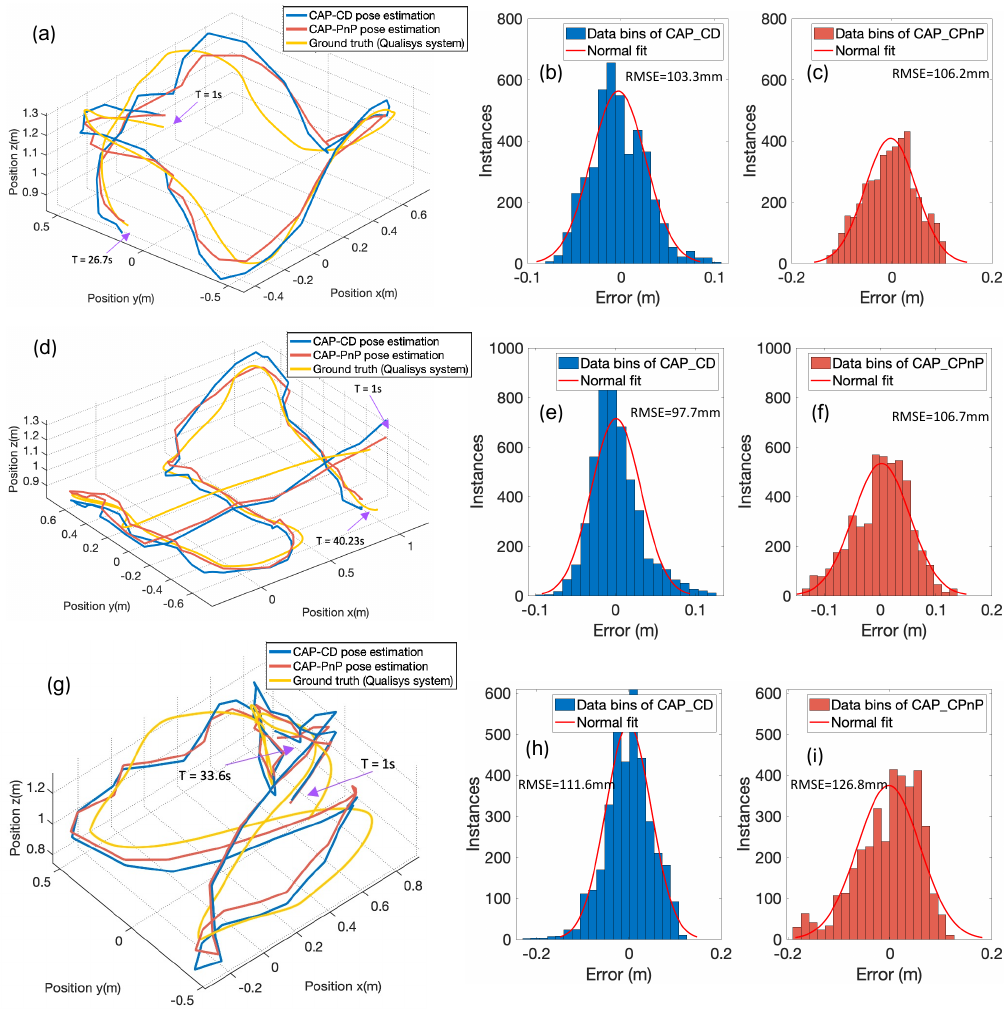}
    \caption{\textbf{The positioning results of the CAP system while the underwater robot moves in different patterns.} Figures(a), (d), and (g) respectively show the trajectory plots of the CAP system positioning underwater robots operating in square, lawnmower, and random patterns. Figures(b), (e), and (h) present the error histograms of CAP-CD for Euclidean distance. Figures(c), (f), and (i) display the error histograms of CAP-CPnP for Euclidean distance.}
    \label{fig:traj_other}
\end{figure*}

Figure~\ref{fig:result1}(a) displays overlaid snapshots of the real-time positioning trajectory of the underwater robot over a duration of 21 seconds. Figure~\ref{fig:result1}(b) shows the positioning performance of CAP\hbox{-}CD and CAP\hbox{-}CPnP, corresponding to the results in Figure~\ref{fig:result1}(a). These overlaid snapshots and trajectories are presented for clarity and conciseness. Beyond the fixed-depth square trajectory illustrated, the performance of CAP\hbox{-}CD and CAP\hbox{-}CPnP under various more complex trajectories is detailed in Figure~\ref{fig:traj_other} and Table~\ref{tab:performance}.
\textcolor{blue}{
While 3D plots are useful for visualising the overall trajectory and spatial context, they can make it difficult to quantify positioning performance at specific moments. Therefore, 2D plots of the Euclidean position error over time are also provided in Appendix~\ref{sec:1D}}.
Figures~\ref{fig:result1}(c), (f), and (i) illustrate the translation of the underwater robot in the world-fixed frame, estimated using both CAP methods in comparison with the ground truth, for the $X$, $Y$, and $Z$ axes respectively.
In Figure~\ref{fig:result1}(d) and (g), it is evident that the positioning error of CAP\nobreakdash-CD increased during changes in the underwater robot's direction of motion, both in the X and Y axes, presented as fluctuations within the graphs.
This phenomenon occurred because CAP\nobreakdash-CD assumed the depth measured by the depth sensor was at the centre of rotation of the underwater robot.
However, the sensor was actually located towards the rear of the robot and changed when the robot's motion caused the robot to tilt.
In contrast, the ground truth, which tracked the centre of the robot, was minimally impacted by this tilt.
The depth sensor, positioned at the BlueROV2's rear, registered significant depth alterations due to the BlueROV2's inclination, which lead to observable positioning fluctuations along the X-Y plane.
Similarly, since the AprilTag was located towards the rear of the robot rather than at the centre, \mbox{CAP-CPnP} experienced the same issue.

% Overall both CAP methods track the ground truth with acceptable accuracy.
%
In terms of the positioning results from the X, Y, and Z axes, the Euclidian root mean square error of CAP\nobreakdash-CD was slightly lower than that of \mbox{CAP-CPnP}, shown in Figure~\ref{fig:result1}(d)(e), (g)(h) and (j)(k).
The MED of CAP\nobreakdash-CD over the 120~s was 70.2~mm, while that of \mbox{CAP-CPnP} was slightly higher at 100.3~mm. (The mathematical definition of MED can be found in Appendix~\ref{sec:MED}.)
Table~\ref{tab:performance} shows the full breakdown of the results for each dataset.
The results indicate that across a variety of trajectories, the accuracy of CAP-CD surpasses that of CAP-CPnP in the X, Y, and Z axes, respectively. Consequently, the MED of CAP-CD is lower than that of CAP-CPnP. The MED for CAP-CD is concentrated between 90~mm to 130~mm, with th
e highest value reaching 123.4~mm.

\textcolor{blue}{
To assess the impact of surface vehicle motion on localisation accuracy, the relationship between the vehicle’s pitch and roll angles and the resulting localisation error was analyzed across the three experimental patterns. Scatter plots of the Euclidean error as a function of pitch and roll are presented in Appendix~\ref{appendix: tilting_accuracy}. 
}

To demonstrate consistency between the results from different experiments, results of the CAP system operating on an underwater robot in square, lawnmower, and random patterns (accompanied by variations in depth) are shown in the Figure~\ref{fig:traj_other}. More details and plots of the results for each dataset and MOVIE can be found in Appendix~\ref{sec:data_code_movies}.

\begin{table*}[ht]
\centering
\caption{Comparative performance of CAP-CD versus CAP-CPnP across various datasets and trajectory types. \textcolor{blue}{The method that performed best for each trajectory and dataset, as well as for each axis, has been bolded.}}
\label{tab:performance}
    \centering
    \includegraphics[width=\textwidth]{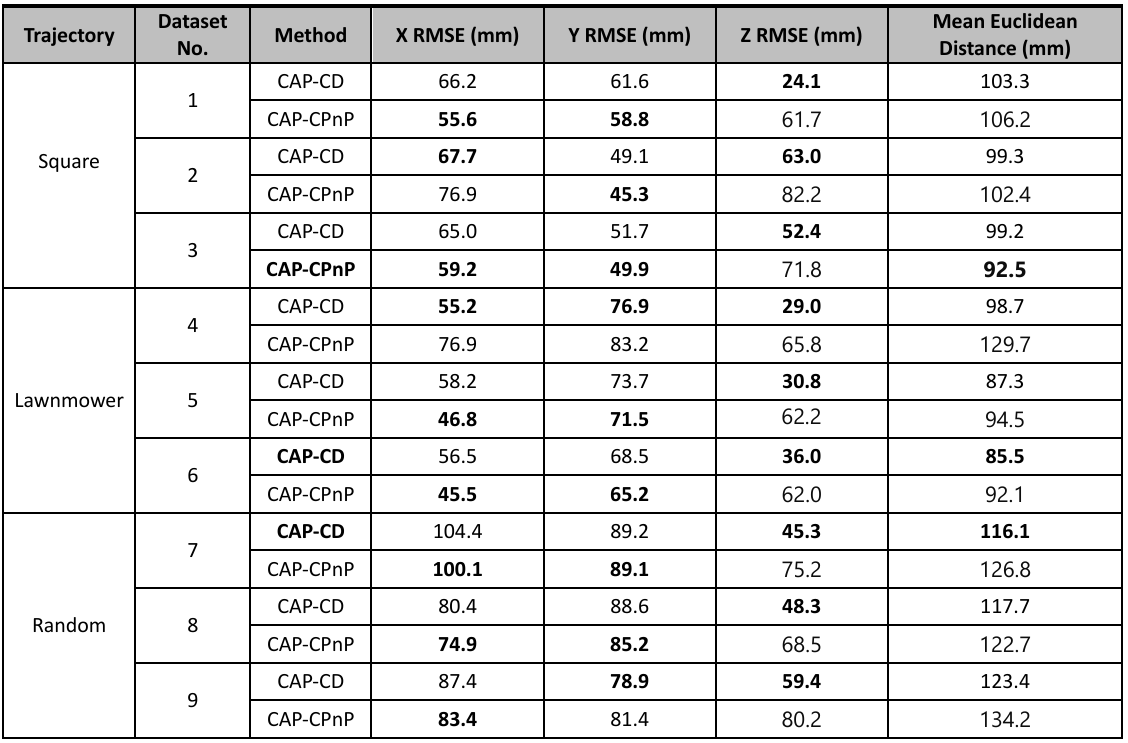}
\end{table*}
% \begin{figure}[ht]
%     \centering
%     \includegraphics[width=\textwidth]{figures/Table 1KG.pdf}
%     \label{fig:table}
% \end{figure} 
\subsection{CAP system in waters of varying turbidity}

The accuracy of positioning in waters with varying levels of turbidity is an important metric for underwater positioning systems.
To address this, a further set of experiments were designed whereby a fiducial marker was laminated and fixed in the middle of the underwater test volume, and the water’s turbidity altered by adding talcum powder.
During the experiments, the water's turbidity was adjusted to 0.12~NTU (Nephelometric Turbidity Units), 2.58~NTU, and 3.74~NTU.
The YSI ProDSS water quality meter was used to measure turbidity and this probe was calibrated using deionized water.
At each turbidity level, three sets of experiments were conducted at different depths: 0.9~m, 1.4~m, and 1.9~m.
Simultaneously, at each depth, the surface robot was programmed to move along three trajectories: a square (with the surface robot performing a 90-degree turn at each corner of the square), a hexagon, and a lawnmower pattern (see MOVIE 2 in Appendix~\ref
{sec:data_code_movies}). The outputs of the CAP system and all sensors were recorded throughout these tests. To refer to the experimental setup, Appendix~\ref{sec:tubidity_setup}.

\begin{figure*}[htbp]
    \centering
    \includegraphics[width=\textwidth]{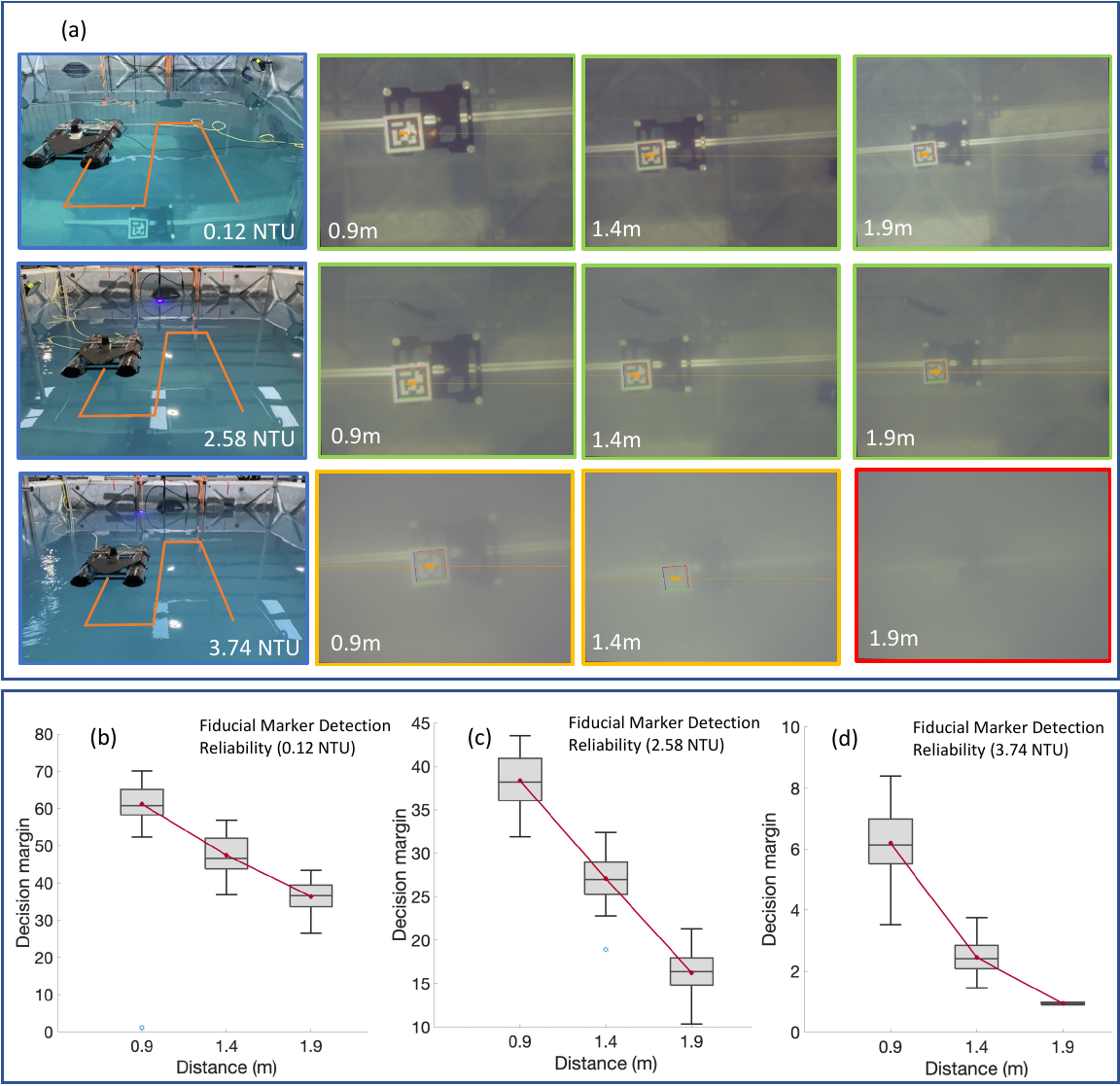}
    \caption{\textbf{CAP system in waters of varying turbidity.}(\textbf{(a)}) In situations where an underwater AprilTag remains stationary while the surface robot moves, and under varying depths as well as different turbidity levels of water, the downward-facing camera's recognition performance of the fiducial marker. \textbf{(b)}, \textbf{(c)} and \textbf{(d)} are the box plots of the decision margins detected by fiducial detection under three different levels of turbidity.}
    \label{fig:turbidity}
\end{figure*}

In these tests, the camera could clearly detect the AprilTag at turbidity levels ranging from 0.12~NTU to 2.58~NTU, as shown in Figure~\ref{fig:turbidity}.
The ability to detect the underwater fiducial marker correlates to two factors: turbidity and the distance between the camera and the target tag.
The confidence in tag detection is quantified by the decision margin. 
\textcolor{blue}{
Decision margin is a scalar value that quantifies the confidence in a tag detection. It typically reflects the difference between the best match score and the second-best match when decoding the tag. A higher decision margin indicates a more confident detection, as the correct tag pattern is clearly distinguishable from other possible candidates.~\cite{decisionmargin}
}
As might be expected, the decision margin demonstrates a negative correlation with both water turbidity and the distance between camera and tag, as shown in Figure~\ref{fig:turbidity}(b), (c) and (d).
% However, when the turbidity rose to 3.74 NTU, the camera's ability to detect the AprilTag was significantly reduced, limiting detection to only a few specific positions, like when the AprilTag was moved very close to the camera.
As turbidity increased to 3.74 NTU, a significant reduction in the decision margin was observed. With turbidity at 3.74 NTU and the distance set to 1.9~m, the efficacy of tag detection diminished considerably, although there were sporadic instances of successful detection under these circumstances.
% When the turbidity rose to 3.74~NTU, there was a noticeable decrease in the decision margin. At a turbidity level of 3.74~NTU and a distance of 1.9~m, tag detection was poor, yet the data showed occasional successful detection in these conditions.

\subsection{Limitations of the study}
 
% A collaborative underwater positioning system, specifically designed to operate in confined environments, was successfully validated in confined indoor experimental tank.
%
A current limitation is that since the underwater portion of the proposed method is vision-based, it inherits some common issues of optical positioning; for instance, ambient lighting and turbidity.
The experimental setup for the turbidity test, conducted within a water tank with a depth of 2.4~m limits the applicability of the results obtained.
To fully assess the capabilities of the CAP system in locating the robot, it would be necessary to conduct measurements in environments that are deeper than the current experimental setup.
However, the turbidity study demonstrates that tag detection has a reasonable degree of tolerance to turbid water.

Although the use of the positioning system as feedback to enable autonomous underwater missions has confirmed that the CAP system can operate successfully in real-time, this does not fully address the issue of temporal synchronization among multiple sensors.
Instead the current system relies on using the most recent update from each sensor.
The current system provides the Cartesian position of the underwater robot in 3-DOF.
However, the system could be expanded to cover 3-DOF rotation with relative ease by re-using components that were developed to determine the surface vehicle's 3-DOF rotation.

An important practical limitation of the setup is that it requires an open expanse of water surface above the underwater environment to allow deployment of the surface vehicle.
In addition, the current system, with only one collaborating robot, can only operate to a depth where the fiducial marker can be tracked. Besides, as the self-localisation of surface robots relies on LiDAR-based SLAM, it necessitates that the environment above the water surface contains features within the range of LiDAR.

\subsection{Discussion}
The positioning of robots in constrained underwater environments introduces a range of challenges that are typically not encountered in open marine environments. Perhaps the most significant of these is that in a constrained environment, an underwater robot may require far greater positional accuracy than would be required in a marine environment. 
Whilst it is possible to design underwater positioning systems for environments where infrastructure, such as cameras or markers, can be added, or where there are a rich set of features, this research has presented a collaborative system that provides underwater positioning in confined or otherwise constrained underwater environments, without the need for infrastructure, system calibration or the environment to be feature rich.
%
%Paragraph about results compared to goals
%
% The goal of this ongoing research is to provide an underwater positioning system that can function in highly physically constrained environments with suitable accuracy to facilitate repeatable and reliable autonomous robotic missions.
%
This capability marks a significant step towards enabling repeatable and reliable autonomous robotic missions in such challenging conditions.
Comparative analysis indicates that the system's performance aligns well with the set objectives, offering a viable solution for the precise navigation and positioning that is required for successful underwater explorations and tasks. 

\textcolor{blue}{
To directly compare CAP-CD and CAP-CPnP focus should be placed on the $X$ and y axes as CAP-CD has the benefit of the depth sensor in $Z$. As shown in Table~\ref{tab:performance}, the CAP-CD and CAP-CPnP methods exhibit comparable performance in the $X$ and $Y$ directions. This is to be expected and is due to the relatively small roll and pitch angles in the conducted experiments. However, as tilting angles increase, error in the camera's z axis will increasingly effect $X$ and $Y$ positions in the world coordinate system. 
Moreover, CAP-CD does not require the use of fiducial markers such as AprilTag. Instead, it only relies on observing a single identifiable point on the underwater robot, which enhances its applicability in marker-free or visually degraded environments. Therefore, despite similar performance in the horizontal axes under mild conditions, CAP-CD offers substantial advantages in flexibility, and extensibility.
}

An important point to note regarding the differences between the two CAP formulations is that, in addition to improved accuracy, the CAP\nobreakdash-CD formulation is more flexible.
The PnP element of the \mbox{CAP-CPnP} formulation relies on four corners of a geometric tag of known dimensions being identified.
This is somewhat limiting, in that a fiducial marker tracking system must generally be employed.
On the other hand, the CAP\nobreakdash-CD system only requires a single point in a projected plane.
This means that the system is open for use with other tracking systems; for instance, those based on deep learning, such as YOLO~\cite{redmon2016you} or fast RCNN~\cite{girshick2015fast} which could track the robot without the need for a fiducial marker.
This would have the added benefit of not requiring the underwater vehicle to be locked in roll and pitch, to maintain visibility of the fiducial marker.
%

%%%%%%%%%%%%%%%%%% CHAPTER 5 %%%%%%%%%%%%%%%%%%
\section{Conclusion and Future Works}\label{sec:dis} 
% \section{CONCLUSION}
% A conclusion section is not required. Although a conclusion may review the
% main points of the paper, do not replicate the abstract as the conclusion. A
% conclusion might elaborate on the importance of the work or suggest
% applications and extensions.
\subsection{Conclusion}
In conclusion, the CAP system proposed in this work provides a robust and balanced solution to the challenges identified in confined underwater environments, taking into account the lack of pre-deployed infrastructure, featureless underwater environments, and limited operating range, while maintaining accuracy.
Furthermore, the proposed novel CAP-CD system also enhances accuracy specifically for underwater robot localisation, enabling positioning with the requirement of knowing only a single feature point of the underwater robot.
It also demonstrates the localisation capability of the CAP system in a replica of a nuclear storage pond, across 9-trajectory datasets.
As a prerequisite for robotic automation, the underwater robot positioning provided by the CAP system can be applied to critical real-world fields, such as automated routine inspection and maintenance of nuclear fuel ponds. It can even be extended to the automation of ship hull and dam inspections.

\subsection{Possible extensions}
 
The CAP positioning system has been designed to be applicable for a broad range of restricted underwater environments, hence the consideration of water turbidity and insufficient ambient lighting conditions.
However, to expand its capabilities further, future work will involve using acoustic sensors, such as a short range  multibeam sonar, to locate the underwater robot rather than the optical cameras. It is anticipated that this approach should enable the positioning system to have improved capability in highly turbid environments.
To extend the capabilities of the proposed system further, it is feasible that multiple underwater vehicles could collaborate, enabling one underwater vehicle, in light of sight of the surface vehicle, to position a second underwater vehicle, not within light of sight of the surface vehicle. This could allow the range, in terms of depth, to be extended, and the system to be used for navigation of highly constrained environments where there is limited, or no line of sight between the submersible robot and surface of the water.
For CAP-CD, currently, the plane in which the underwater robot, as defined by the depth sensor, resides does not take into account the pitch and roll rotations of the underwater robot. Once rotation occurs, the plane of the robot's depth will not be equivalent to the plane in which the depth sensor is located. To address this, the pose of the underwater robot can be obtained through an IMU, thereby acquiring accurate depth information.

\textcolor{blue}{
 The current single underwater robot framework could be extended to include more underwater robots arranged in a chain configuration, facilitating a deeper coveragge zone and the possibility of operating underneath surface infrastructure. However error propagation would need to be carefully considered. The chain of robots can be modelled analogously to a multi-link manipulator, where each inter-robot observation acts like a revolute or prismatic joint. error would propagate through a chain of inter-robot observations via Jacobian-based covariance propagation~\cite{wang2006error}, as in a manipulator's forward kinematics, leading to a growth in uncertainty with respect to the number of robots.
 }

\textcolor{blue}{
Another potential extension is to replace fiducial marker detection with deep learning-based object detection methods such as YOLO. However, this introduces several challenges. Unlike fiducial systems, YOLO-based detection lacks sub-pixel accuracy and does not provide 6-DoF pose information. Additionally, the reliability of YOLO deteriorates under non-frontal or tilted views, which are common in underwater scenarios. These limitations would likely lead to discontinuities in detection and reduced localisation precision. Therefore, while deep learning-based methods offer flexibility, fiducial markers remain more suitable for high-accuracy pose estimation in this system.
}

\textcolor{green}{In real nuclear pond inspection scenarios, system safety and redundancy are of critical importance. Although the current study primarily focuses on demonstrating the feasibility and accuracy of the CAP-CD system, future research will place emphasis on robustness and fail-safe operation. When a localisation outage occurs between the surface robot and the underwater robot, the underwater vehicle can rely on onboard sensors such as an IMU, DVL, or even a camera to perform short-term dead reckoning for pose estimation until the CAP system reconnects.}
 % 
 % In high-dimensional rotational chains, this growth can resemble exponential accumulation due to nonlinearity and compounding uncertainty.
%
% The error propagation through a chain of inter-robot observations can be described using a Jacobian-based covariance propagation function, similar to forward kinematics in a manipulator:
% $$
% \Sigma_{\text {final }}=J \Sigma J^{\top},
% $$
% where:
% \begin{itemize}
%     \item $\Sigma_{\text {final }}$ : the covariance (uncertainty) of the final robot's position and orientation,
%     \item $J$ : the Jacobian matrix mapping uncertainties from intermediate robots to the final robot,
%     \item $\Sigma$ : the covariance of each intermediate step (robot) or relative observation.
% \end{itemize}
% As the number of robots (analogous to the number of joints in a manipulator) increases, the accumulated uncertainty can exhibit near-exponential growth, particularly in systems involving rotational transformations.

\section*{Acknowledgments}
This work was supported by Robotics and Artificial Intelligence Collaboration (RAICo) lab between the UK Atomic Energy Authority (UKAEA), Nuclear Decommissioning Authority (NDA), Sellafield Ltd and the University of Manchester, through Engineering and Physical Sciences Research Council (EPSRC) grants: EP/P01366X/1, EP/W001128/1 and EP/V026941/1.

\bibliographystyle{ieeetr}
\bibliography{07_reference}

\newpage
\section{Biography Section}
 

% \end{IEEEbiography}



\begin{IEEEbiographynophoto}{XUELIANG CHENG}, Robotics for Extreme Environments Group, Robotics and AI, Department of Electric and Electronic Enginnering, the University of Manchester. Email: xueliang.cheng@manchester.ac.uk
\end{IEEEbiographynophoto}

\begin{IEEEbiographynophoto}{KANZHONG YAO}, Robotics for Extreme Environments Group, Robotics and AI, Department of Electric and Electronic Enginnering, the University of Manchester. Email: xueliang.cheng@manchester.ac.uk
\end{IEEEbiographynophoto}

\begin{IEEEbiographynophoto}{ANDREW WEST}, Robotics for Extreme Environments Group, Robotics and AI, Department of Electric and Electronic Enginnering, the University of Manchester. Email: xueliang.cheng@manchester.ac.uk
\end{IEEEbiographynophoto}

\begin{IEEEbiographynophoto}{SIMON WASTON}, Robotics for Extreme Environments Group, Robotics and AI, Department of Electric and Electronic Enginnering, the University of Manchester. Email: xueliang.cheng@manchester.ac.uk
\end{IEEEbiographynophoto}

\begin{IEEEbiographynophoto}{OGNJEN MARJANOVIC}, Robotics for Extreme Environments Group, Robotics and AI, Department of Electric and Electronic Enginnering, the University of Manchester. Email: xueliang.cheng@manchester.ac.uk
\end{IEEEbiographynophoto}

\begin{IEEEbiographynophoto}{BARRY LENNOX}, Robotics for Extreme Environments Group, Robotics and AI, Department of Electric and Electronic Enginnering, the University of Manchester. Email: xueliang.cheng@manchester.ac.uk
\end{IEEEbiographynophoto}

\begin{IEEEbiographynophoto}{KEIR GROVES}, Robotics for Extreme Environments Group, Robotics and AI, Department of Electric and Electronic Enginnering, the University of Manchester. Email: xueliang.cheng@manchester.ac.uk
\end{IEEEbiographynophoto}

\vfill

\newpage
\appendix
\setcounter{subsection}{0}
\subsection{Source code, data and movies}\label{sec:data_code_movies}

\begin{itemize}
    \item Datasets are avilable at IEEE DataPort: \\
    DOI:10.21227/6z0y-yf36
    \item The source code of CAP system, datasets and movies can also be found in: \\
\url{https://livemanchesterac-my.sharepoint.com/:f:/g/personal/xueliang_cheng_postgrad_manchester_ac_uk/Em_0ol2h60dLicXANMnqVfQB1TRxVJHSAZMBNJCb7zPBfg?e=XcTmyB} 

    \item Movie is also available in: \\
    \url{https://www.youtube.com/watch?v=rkGgttDFZlw}
\end{itemize}

\subsection{Calculating ASV rotation relative to the world frame}
\label{sec:imu_align}
In order to derive the position of the marker in the world coordinate system, as Equation~(\ref{equ:2homos}), it is required to compute a homogeneous transformation $\mathbf{H}^W_B$, which comprises $\mathbf{R}^W_B$ and $\mathbf{p}^W_B$. 
\begin{equation}
\mathbf{P}^W_C=\mathbf{H}^W_B \mathbf{P}_C^B
% \label{equ:2homos}
\end{equation}

\begin{figure}[ht]
    \centering
    \includegraphics[width=0.9\columnwidth]{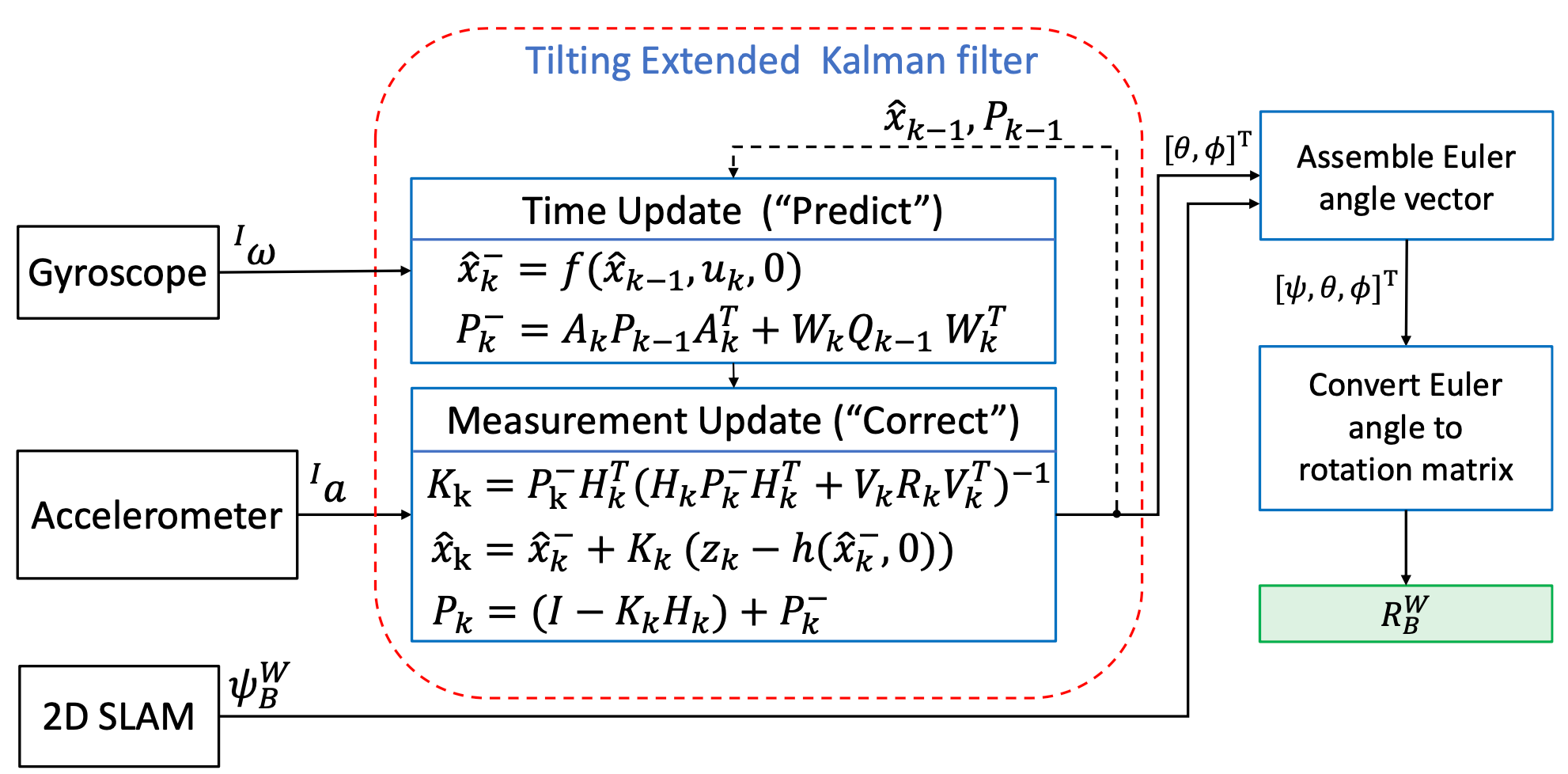}
    \caption{Overview of the proposed EKF structure. The superscripts + and - stand for the 'a posteriori' and the 'a priori' estimates in the Kalman filter}
    \label{fig:2stages}
\end{figure}

In this case, an extended Kalman filter was employed to fuse IMU and SLAM data for orientation of the USV relative to the world frame $\mathbf{R}^W_B$. Given that we will obtain the tilting angles from the IMU and the yaw angle from SLAM, subsequently fusing them into a comprehensive three-dimensional rotation, we find it convenient to use the Euler angle sequence of Z-Y-X. While the issue of gimbal lock is a known problem with Euler angles, it will not occur in this case. For surface vehicles, when rotating in the Z-Y-X sequence, the rotation around the Y-axis will not reach 90 degrees, effectively preventing gimbal lock. The tilting EKF has two inputs, as shown in Fig\,\ref{fig:2stages}: 3-axis angular rate $\boldsymbol{\omega} = [{\omega}_x, {\omega}_y, {\omega}_z]^{\top}$ measured by gyroscope(G), 3-axis acceleration  $\boldsymbol{a} = [{a}_x, {a}_y, {a}_z]^{\top}$ measured by accelerometer in IMU sensor frame, 
where the gyro and accelerometer measurement noise are assumed to be uncorrelated Gaussian noise. In this study, the motion of the USV does not exhibit prolonged substantial additional accelerations (other than gravitational acceleration) for an extended period of time. Hence, any additional linear acceleration is assumed to be non-existent or effectively zero in this context.

% The study focuses on the motion of the Unmanned Surface Vehicle (USV) where it doesn't experience significant or sustained additional accelerations for a prolonged duration beyond what is provided by the force of gravity. As such, any additional linear acceleration is assumed to be non-existent or effectively zero in this context.

% In this project, a low-pass filter can effectively distinguish between gravity acceleration and other additional accelerations for the following reasons: 1) Gravity acceleration is typically a constant, low-frequency signal. The acceleration due to gravity at Earth's surface is approximately 9.81 m/s². For stationary or slowly moving objects, the gravity acceleration is a relatively stable value. Thus, the frequency components of gravity acceleration are low. 2) Additional accelerations, such as those generated by the dynamic motion of a robot, usually manifest as high-frequency signals. This is because the robot may encounter rapidly changing forces during its movement, such as collisions and vibrations, which often result in accelerations that vary quickly in both time and frequency.

\subsection{\textbf{Tilting Extended Kalman filter (pitch and roll)}}\label{appendix:tkf}
The extended Kalman filter for tilt estimation facilitates precise computation of Euler angle roll ($\phi$) and pitch ($\theta$) in dynamically varying conditions. In this approach, data from tri-axial gyroscope and accelerometer are incorporated within a Kalman filter to approximate the normalized gravity vector within the sensor frame, utilizing the succeeding system model equations:

\begin{equation}
x_t=f\left(x_{t-1}, \omega_{t-1}\right)
\label{eq:prediction}
\end{equation}

\begin{equation}
z_t=h(x_t,v_t)
\label{eq:observation}
\end{equation}

In equation~\ref{eq:prediction} and \ref{eq:observation}, $\boldsymbol{x}_t=\left[\phi, \theta \right]^T$ is 2 $\times$ 1 state vector at step k. $\boldsymbol{\omega_t}$ is process noise, and $\boldsymbol{v_t}$ is observation noise. The prediction without the noise can be calculated as follows:

\begin{equation}
\begin{gathered}
\Tilde{x}_t=f\left(\hat{x}_{t-1}, 0\right)=\left[\begin{array}{c}
\phi+\dot{\phi} \Delta t \\
\theta+\dot{\theta} \Delta t
\end{array}\right] \\
=\left[\begin{array}{c}
\phi+\Delta t\left(\omega_x+\omega_y \cdot s\phi \cdot t\theta+\omega_z \cdot c\phi \cdot t\theta\right) \\
\theta+\Delta t\left(\omega_y \cdot c\phi-\omega_z \cdot s\phi\right)
\end{array}\right],
\end{gathered}
\end{equation}
where $c \psi$ is $cos\psi$, $s \psi$ is $sin\psi$ and $t\theta$ is $\tan\theta$. $\Delta t$ represents time step size. $\omega_x$, $\omega_y$ and $\omega_z$ are tri-axis gyroscope measurements. The state observation in (\ref{eq:observation}) is calculated by:

\begin{equation}
\tilde{z}_k=h\left(\tilde{x}_k, 0\right)=\left[\begin{array}{l}
\phi \\
\theta
\end{array}\right]=\left[\begin{array}{c}
\tan ^{-1}\left(\frac{a_x}{\sqrt{a_y^2+a_z^2}}\right) \\
\tan ^{-1}\left(\frac{-a_y}{-a_z}\right)
\end{array}\right],
\label{eq:z}
\end{equation}
where the $a_x, a_y$ and $a_z$ in (\,\ref{eq:z}) are the tri-axis acceleration measured by the accelerometer. The Jacobian matrix of partial derivatives of $f$ with respect to $\mathbf{x}$ is:

% \begin{equation}
% \begin{gathered}
% A_t=\left.\frac{\partial f_{t-1}(x)}{\partial x}\right|_{x=\hat{x}_{t-1}} \\
% =\left[\begin{array}{cc}
% \frac{\partial(\phi+\dot{\phi} \Delta t)}{\partial \phi} & \frac{\partial(\phi+\dot{\phi} \Delta t)}{\partial \theta} \\
% \frac{\partial(\theta+\dot{\theta} \Delta t)}{\partial \phi} & \frac{\partial(\theta+\dot{\theta} \Delta t)}{\partial \theta}
% \end{array}\right] \\
% =\left[\begin{array}{cc}
% 1+\Delta t\left(\omega_y \cdot c \phi \cdot t \theta-\omega_z \cdot s \phi \cdot t \theta\right) & \Delta k\left(\frac{\omega_y \cdot s \phi}{c^2 \theta}+\frac{\omega_z \cdot c \phi}{c^2 \theta}\right) \\
% -\Delta t\left(\omega_y \cdot s \phi+\omega_z \cdot c \phi\right) & 1
% \end{array}\right].
% \end{gathered}
% \end{equation}

\begin{equation}
\resizebox{\columnwidth}{!}{$
\begin{gathered}
A_t=\left.\frac{\partial f_{t-1}(x)}{\partial x}\right|_{x=\hat{x}_{t-1}} \\
=\left[\begin{array}{cc}
\frac{\partial(\phi+\dot{\phi} \Delta t)}{\partial \phi} & \frac{\partial(\phi+\dot{\phi} \Delta t)}{\partial \theta} \\
\frac{\partial(\theta+\dot{\theta} \Delta t)}{\partial \phi} & \frac{\partial(\theta+\dot{\theta} \Delta t)}{\partial \theta}
\end{array}\right] \\
=\left[\begin{array}{cc}
1+\Delta t\left(\omega_y \cdot c \phi \cdot t \theta-\omega_z \cdot s \phi \cdot t \theta\right) & \Delta k\left(\frac{\omega_y \cdot s \phi}{c^2 \theta}+\frac{\omega_z \cdot c \phi}{c^2 \theta}\right) \\
-\Delta t\left(\omega_y \cdot s \phi+\omega_z \cdot c \phi\right) & 1
\end{array}\right].
\end{gathered}
$}
\end{equation}

As illustrated in Fig\,\ref{fig:2stages}, the purpose of the tilting extended Kalman filter is to estimate pitch($\theta$) and roll($\phi$) and will be used to combine with yaw angle computed from 2D SLAM.

As illustrated in Fig\,\ref{fig:2stages}, the output of SLAM and IMU fusion
is Euler angle $[\psi^W_B, \theta^W_B, \phi^W_B]^{\top}$, which pitch($\theta$) and roll($\phi$) are calculated from tilting EKF, the $\boldsymbol{R_B^W}$ can be expressed as the conventional Z-Y-X Euler angles using the Euler angle computed from the fusion:

\begin{equation}
\mathbf{R_B^W}
=\left[\begin{array}{ccc}
c \psi c \theta & c \psi s \theta s \phi-s \psi c \phi & c \psi s \theta c \phi+s \psi s \phi \\
s \psi c \theta & s \psi s \theta s \phi+c \psi c \phi & s \psi s \theta c \phi-c \psi s \phi \\
-s \theta & c \theta s \phi & c \theta c \phi
\end{array}\right]
\end{equation}

\subsection{\textbf{SLAM Yaw angle and IMU tilting angle fusion}}
 In this step, yaw in SLAM map frame, along with the known tilting angles in previous EKF, are employed to estimate the corresponding rotation matrix $\mathbf{R}^W_B$. In our experimental conditions, due to the relatively simple structure of the test tank, the $[x^W_B, y^W_B, \psi^W_B]^{\top}$ provided by SLAM is comparatively accurate, which has been verified in our previous research. 

\subsection{Camera calibration}\label{appdex:camaera_cali}

\begin{figure}[htbp]
    \centering
    \includegraphics[width=\columnwidth]{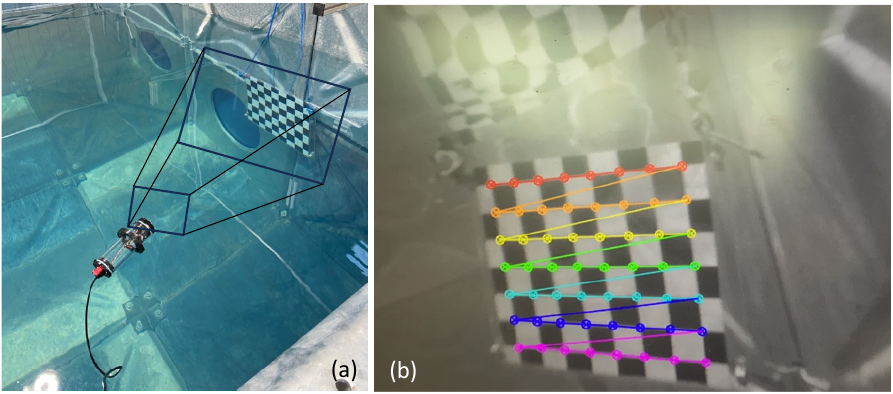}
    \caption{\textbf{Camera calibration underwater using chessboard.} \textbf{(a)}: Camera enclosure underwater. \textbf{(b)}: Grayscale image of a chessboard. The chessboard is correctly registered, with adjacent vertices of each square connected by colored lines.}
    \label{fig:camera_cali}
\end{figure}
To obtain the camera's intrinsic parameters and distortion coefficients, underwater calibration was performed. The calibration followed Zhang's method~\cite{zhang2000flexible}, which was integrated into OpenCV and can be implemented in ROS. 
In the calibration process, the camera was placed inside a waterproof enclosure, and the approach not only accounted for the distortion of the camera lens itself but also for the barrel distortion caused by the enclosure's dome-shaped cover.
\begin{itemize}
    \item Image Width: 800
    \item Image Height: 600
\end{itemize}

\textbf{Camera Intrinsic Matrix:}
\[
\begin{bmatrix}
    514.177765 & 0 & 346.861136 \\
    0 & 513.054629 & 220.015799 \\
    0 & 0 & 1
\end{bmatrix}
\]

\textbf{Distortion Coefficients:}
% \[
% \begin{bmatrix}
%     0.073902 & -0.032694 & -0.001420 & -0.002268 & 0.000000
% \end{bmatrix}
% \]

\begin{adjustbox}{max width=\columnwidth, , center}

[0.073902 -0.032694 -0.001420 -0.002268 0.000000]

\end{adjustbox}

\textbf{Rectification Matrix:}
\[
\begin{bmatrix}
    1 & 0 & 0 \\
    0 & 1 & 0 \\
    0 & 0 & 1
\end{bmatrix}
\]

\textbf{Projection Matrix:}
\[
\begin{bmatrix}
    529.887695 & 0.000000 & 344.207956 & 0.000000 \\
    0.000000 & 530.503540 & 219.071616 & 0.000000 \\
    0.000000 & 0.000000 & 1.000000 & 0.000000
\end{bmatrix}
\]

\subsection{Depth/Pressure Sensor Calibration}\label{appendix:depth_cali}
% \subsection*{Depth sensor calibration}
In the aforementioned CAP\nobreakdash-CD system, the depth sensor plays a crucial role, thus necessitating proper calibration once prior to usage. The calibration process involved leveraging the Qualisys system, known for its high positional accuracy, by attaching the sensor to an ROV submerged in water. This setup facilitated the acquisition of both sensor data and precise ground truth. Calibration was accomplished via Particle Swarm Optimization, correlating sensor-measured depth values to their ground truth counterparts. The cost function for calibrating the depth sensor is defined as the sum of squared differences between the measured depth from the sensor and the actual depth. Mathematically, it is formulated as:

\begin{equation}
f(\theta)=\sum_{i=1}^n\left(A_i \cdot \theta_1+\theta_2-B_i\right)^2
\end{equation}

where: $A$ and $B$ correspond to the depth sensor's measurement and true depth values, respectively. $\theta=\left[\theta_1, \theta_2\right]$ are the calibration parameters, with $\theta_1$ being a scaling factor and $\theta_2$ an offset. $n$ is the number of data points.

First to initialization the PSO, a swarm of particles is created, each representing a potential solution $\theta_j$ for the calibration parameters. Each particle has an initial velocity $v_j$.
Second, the cost function $f\left(\theta_j\right)$ is evaluated for each particle to assess its performance. After that, update the velocity and position of each particle based on its best-known position and the swarm's best-known position. The update rules are:
\begin{equation}
v_j^{(\text{new})} = w \cdot v_j^{(\text{old})} + c_1 \cdot r_1 \cdot (p_{\text{best}, j} - \theta_j) + c_2 \cdot r_2 \cdot (g_{\text{best}} - \theta_j)
\end{equation}

\begin{equation}
    \theta_j^{(\text{new})} = \theta_j^{(\text{old})} + v_j^{(\text{new})},
\end{equation}
where $w$ is the inertia weight, $c_1$ and $c_2$ are the cognitive and social coefficients, respectively, and $r_1, r_2$ are random factors. Through this iterative process, the PSO algorithm aims to find the calibration parameters $\theta$ that minimize the cost function, thereby calibrating the depth sensor effectively.

\begin{figure}[ht]
    \centering
    \includegraphics[width=\columnwidth]{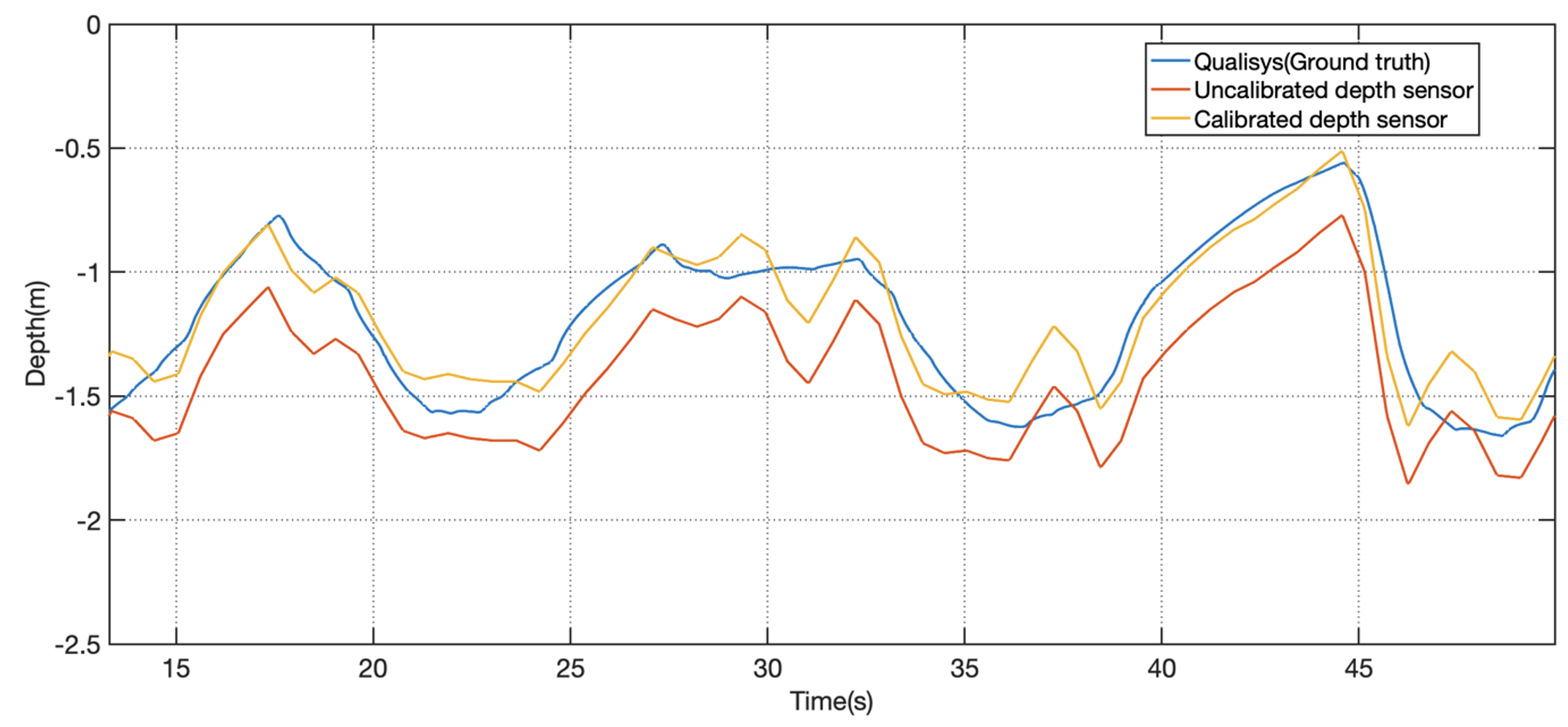}
    \caption{The figure illustrates the depth from Qualisys (blue), the depth sensor's measurement (red), and the calibrated depth sensor (yellow).}
    \label{fig:depth_cali}
\end{figure}

During the calibration of the depth sensor, the ROV performed vertical reciprocating movements within the tank, with an amplitude of approximately 1 meter. Both Qualisys and the depth sensor were simultaneously recorded for calibration purposes. The calibration result is depicted in Figure~\ref{fig:depth_cali}.

\subsection{2D plots}~\label{sec:1D}

\begin{figure}[ht]
    \centering
    \includegraphics[width=\linewidth]{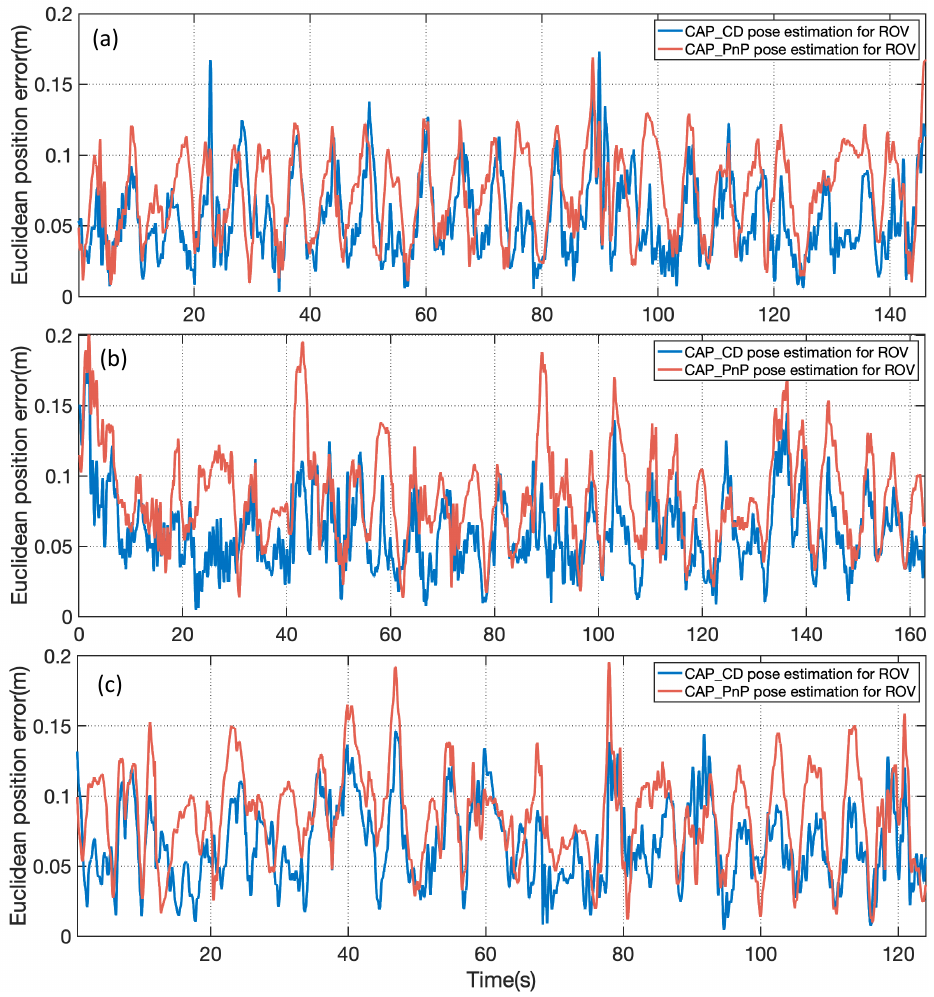}
    \caption{\textcolor{blue}{Time-series plot of the Euclidean position error for the CAP-CD and CAP-CPnP methods. Subfigures (a), (b), and (c) correspond to three different datasets (Dataset 3, Dataset 6, Dataset 9), each representing a distinct experimental scenario.}}
    \label{fig:1D}
\end{figure}

\textcolor{green}{\subsection{Percentage of time fiducial marker successfully tracked in autonomous following tests}}

\begin{table}[H]
    \centering
    \caption{Percentage of time fiducial marker successfully tracked in autonomous following tests}
    \includegraphics[width=\linewidth]{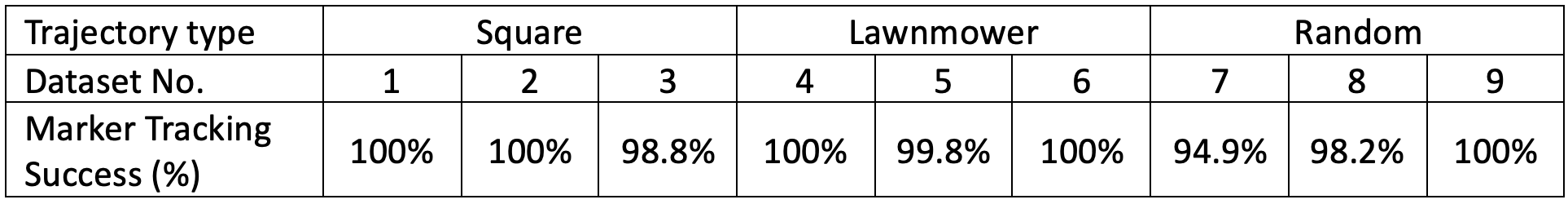}
    \label{tab:marktrackingsuccess}
\end{table}

\subsection{Mean Euclidean distance}\label{sec:MED}
When evaluating the Euclidean distance error, the MED between the output of the CAP system and the ground truth was chosen to assess the overall accuracy of the CAP system.
%
% In this work, the deviations of pose vector components from their respective ground truth values are considered estimation errors. 
%
The formulation of MED is given as follows:

\begin{equation}
\Delta p_{e_i} = \sqrt{ (p_{x_{o_i}}-p_{x_{g_i}})^2 + (p_{y_{o_i}} - p_{y_{g_i}})^2 + (p_{z_{o_i}} - p_{z_{g_i}})^2}
\end{equation}

where $p_{x_{o_i}}$, $p_{y_{o_i}}$ and $p_{z_{o_i}}$ represent the component of pose vector $\boldsymbol{p}$, which is plotted along the \textit{x}-axis, \textit{y}-axis and \textit{z}-axis at timestamp $i$, in the following graphs, along with its estimate, and $p_{x_{g_i}}$, $p_{y_{g_i}}$ and $p_{z_{g_i}}$ correspond to the ground values at timestamp $i$. 

The mean of the Euclidean distance for $n$ estimation steps is given by:
\begin{equation}\label{eq:rmse}
\mu_{\Delta {p_e}} = \frac{\sum_{i=1}^n \Delta {p_{e_i}}}{n}
\end{equation}

\subsection{Turbidity test setup}\label{sec:tubidity_setup}
\begin{figure}[t]
    \centering
    \includegraphics[width=\columnwidth]{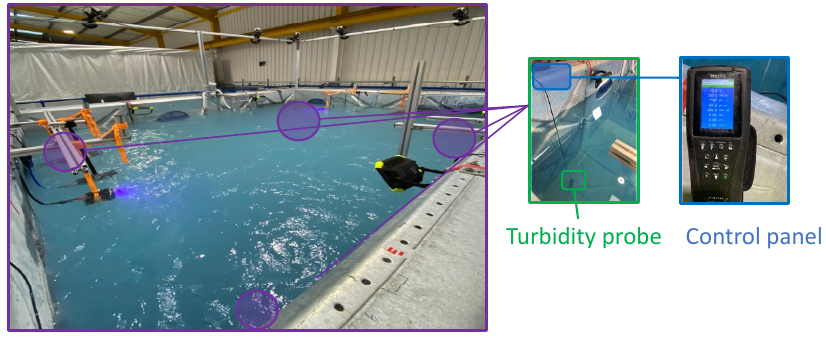}
    \caption{\textbf{Turbidity Test Experimental Setup}}
    \label{fig:turbidity_test_setup}
\end{figure}

When talcum powder was added to change the turbidity, a pump was continuously operated
to keep the talcum powder suspended in the water rather than settling at the bottom. The
turbidity in the tank was measured by placing a turbidity probe at four different locations
within the experimental tank (shown in Fig~\ref{fig:turbidity_test_setup}) and then averaging the results.

% \section{Diagram of the CAP System IP Connection}
% \begin{figure}[htbp]
%     \centering
%     \includegraphics[width=\columnwidth]{figures/IP.png}
%     \caption{IP connection}
%     \label{fig:ip}
% \end{figure}
\clearpage

\subsection{Tilting angles \& positioning accuracy}~\label{appendix: tilting_accuracy}

\begin{figure}[ht]
    \centering
    \includegraphics[width=\linewidth]{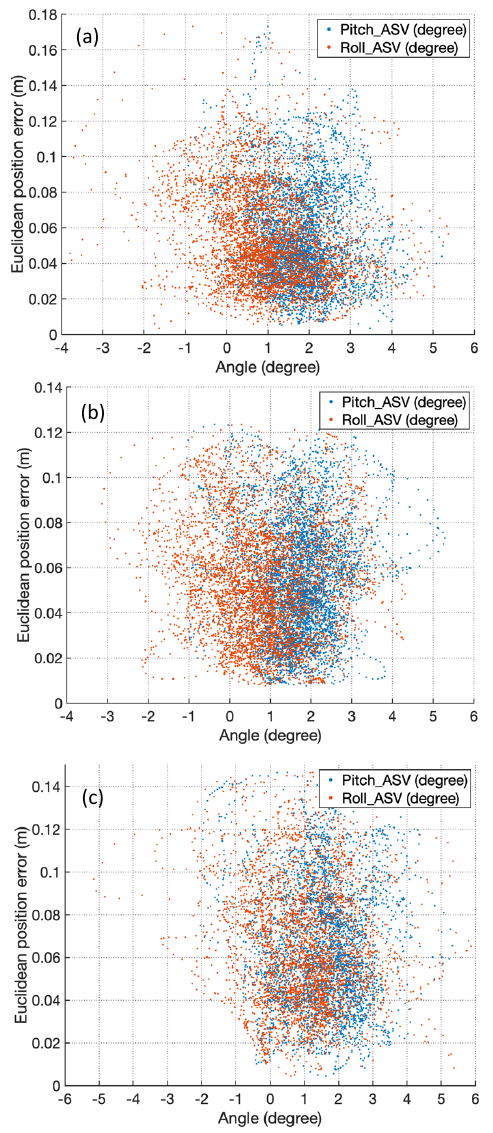}
    \caption{\textcolor{blue}{Scatter plots of Euclidean localisation error versus pitch and roll angles of the surface vehicle, across the three tested motion patterns. Subfigures (a), (b), and (c) correspond to the three different experimental motion patterns.}}
    \label{fig:rollpitcherror}
\end{figure}
\textcolor{blue}{
A linear regression analysis was performed on one of the datasets to evaluate the impact of the surface robot’s pitch and roll on localisation accuracy. The results showed that although both variables were statistically significant, the overall explanatory power of the model was limited (R² = 0.0222). This suggests that pitch and roll do not exhibit a strong linear relationship with the localisation error. 
It indicates no strong linear correlation between pitch or roll and localisation error. These results suggest that the proposed localisation system is relatively independent of moderate surface vehicle maneuvers, while other factors, such as observation geometry or environmental disturbances, may have a more substantial influence on error behaviour.
}

\begin{table}[ht]
    \centering
    \caption{\textcolor{blue}{Linear Regression Coefficients Relating Pitch and Roll to Localisation Error}}
    \includegraphics[width=\linewidth]{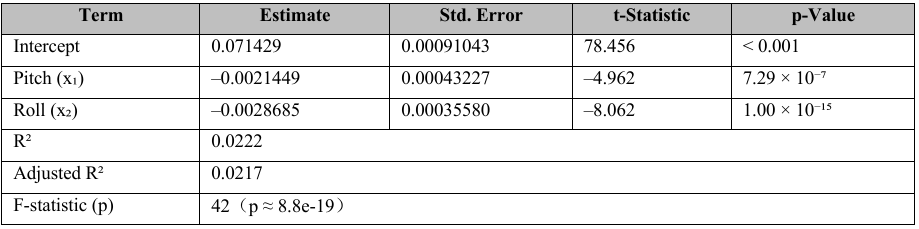}
    \label{tab:my_label}
\end{table}
\clearpage

\subsection{Results of all tests}

\begin{figure*}[ht]
    \centering
    \includegraphics[width=0.9\textwidth]{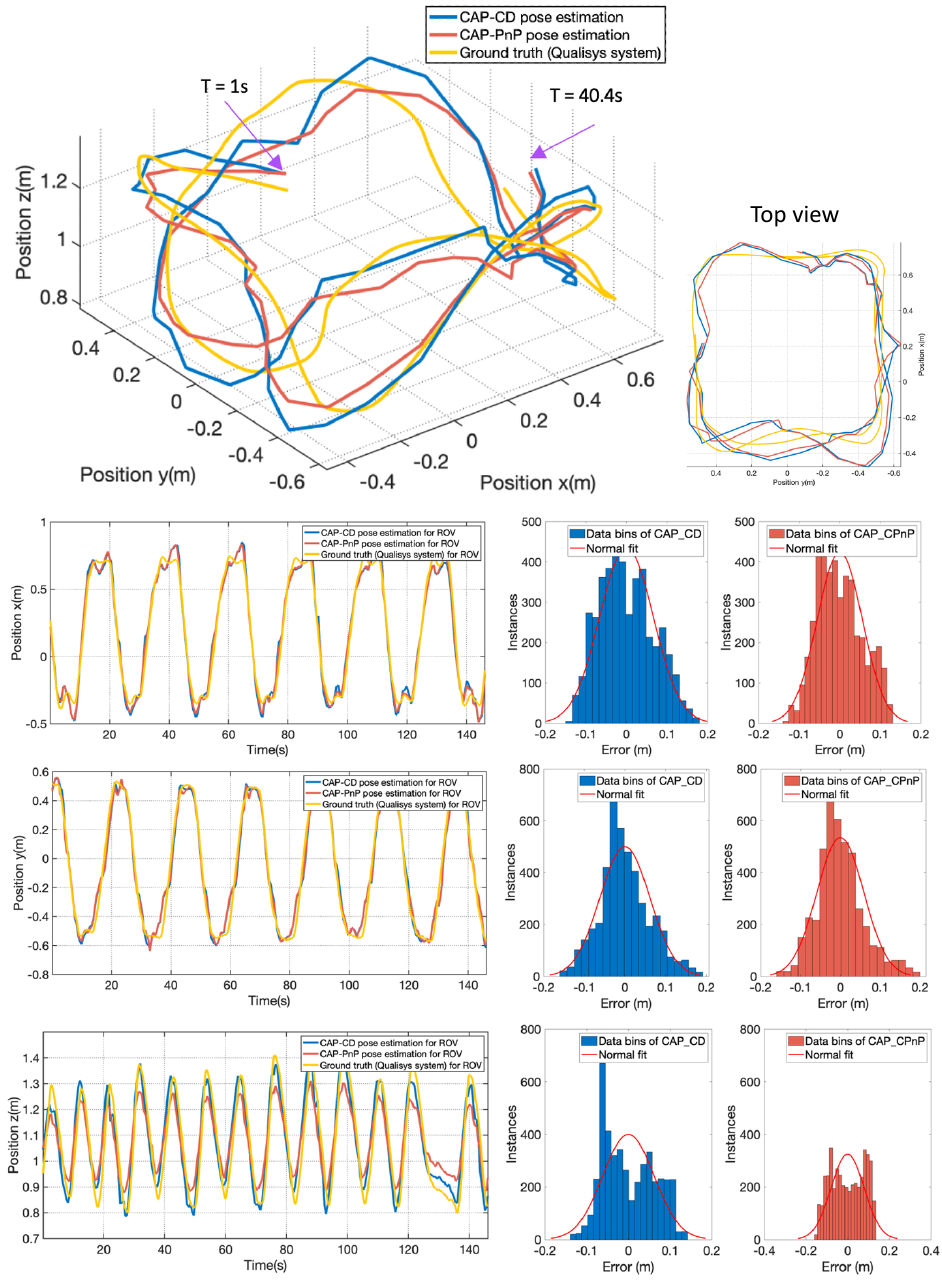}
    \caption{Square pattern 1}
    \label{fig:s1}
\end{figure*}
\clearpage

\begin{figure}[ht]
    \centering
    \includegraphics[width=0.9\textwidth]{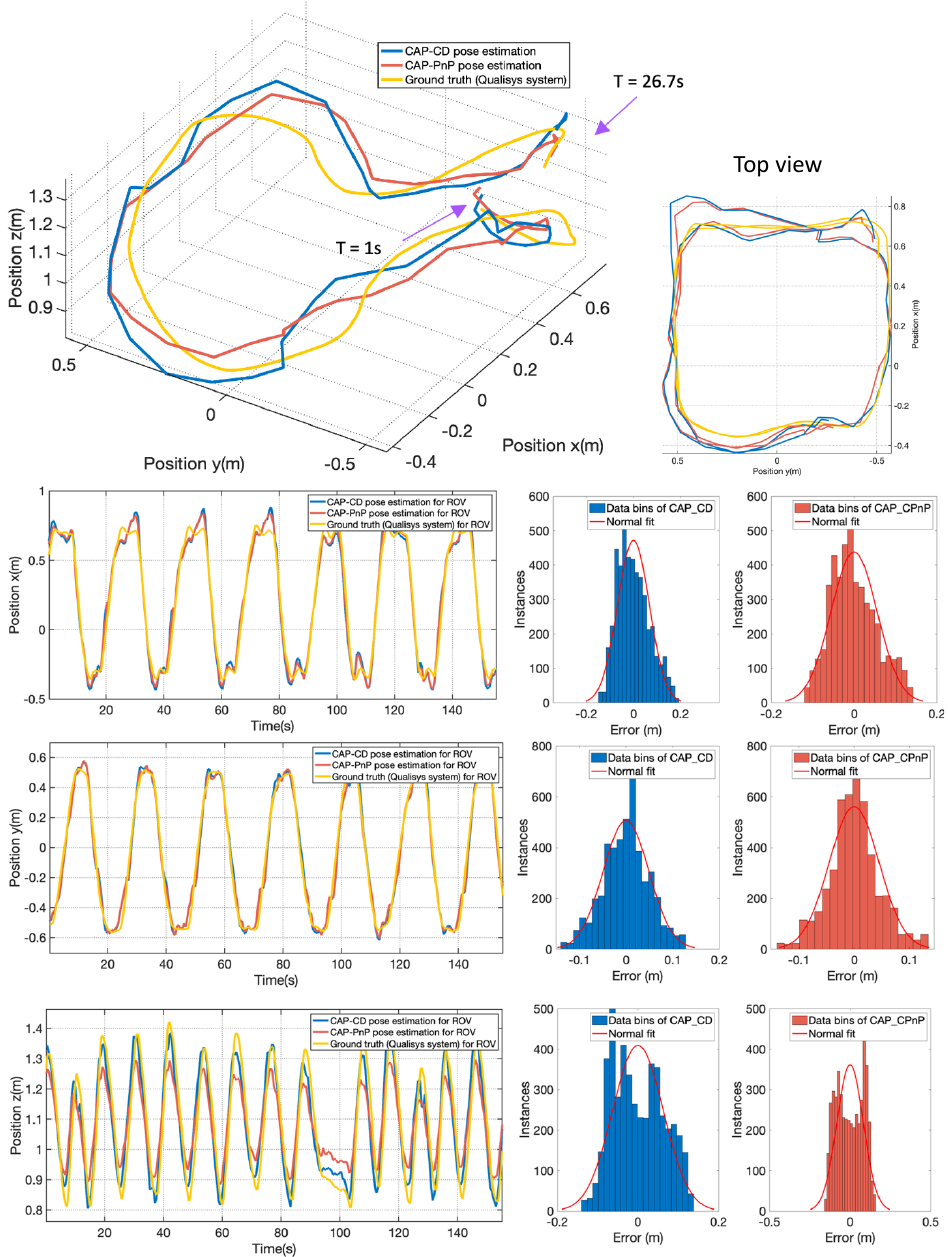}
    \caption{Square pattern 2}
    \label{fig:s2}
\end{figure}
\clearpage

\begin{figure}[ht]
    \centering
    \includegraphics[width=0.9\textwidth]{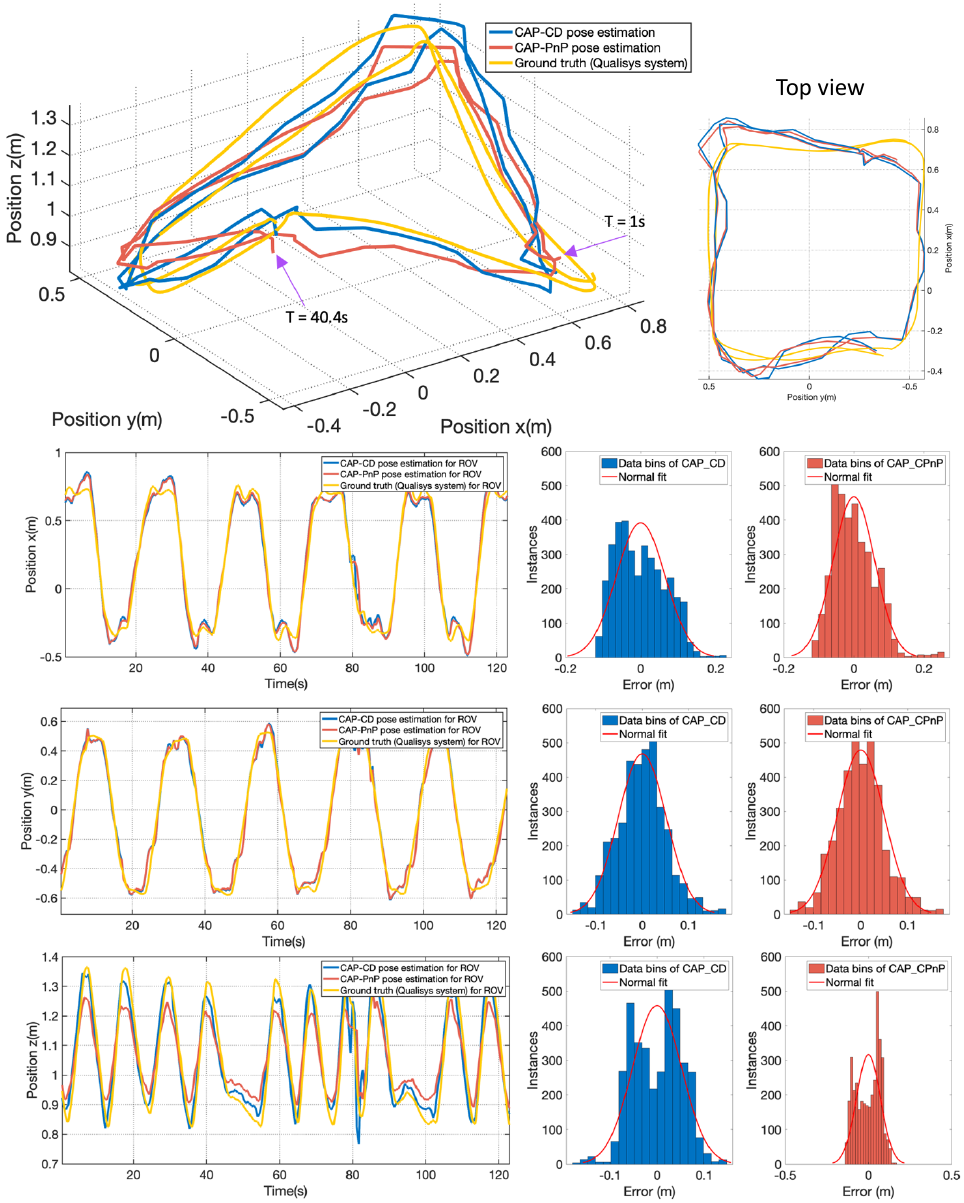}
    \caption{Square pattern 3}
    \label{fig:s3}
\end{figure}
\clearpage

\begin{figure}[ht]
    \centering
    \includegraphics[width=0.9\textwidth]{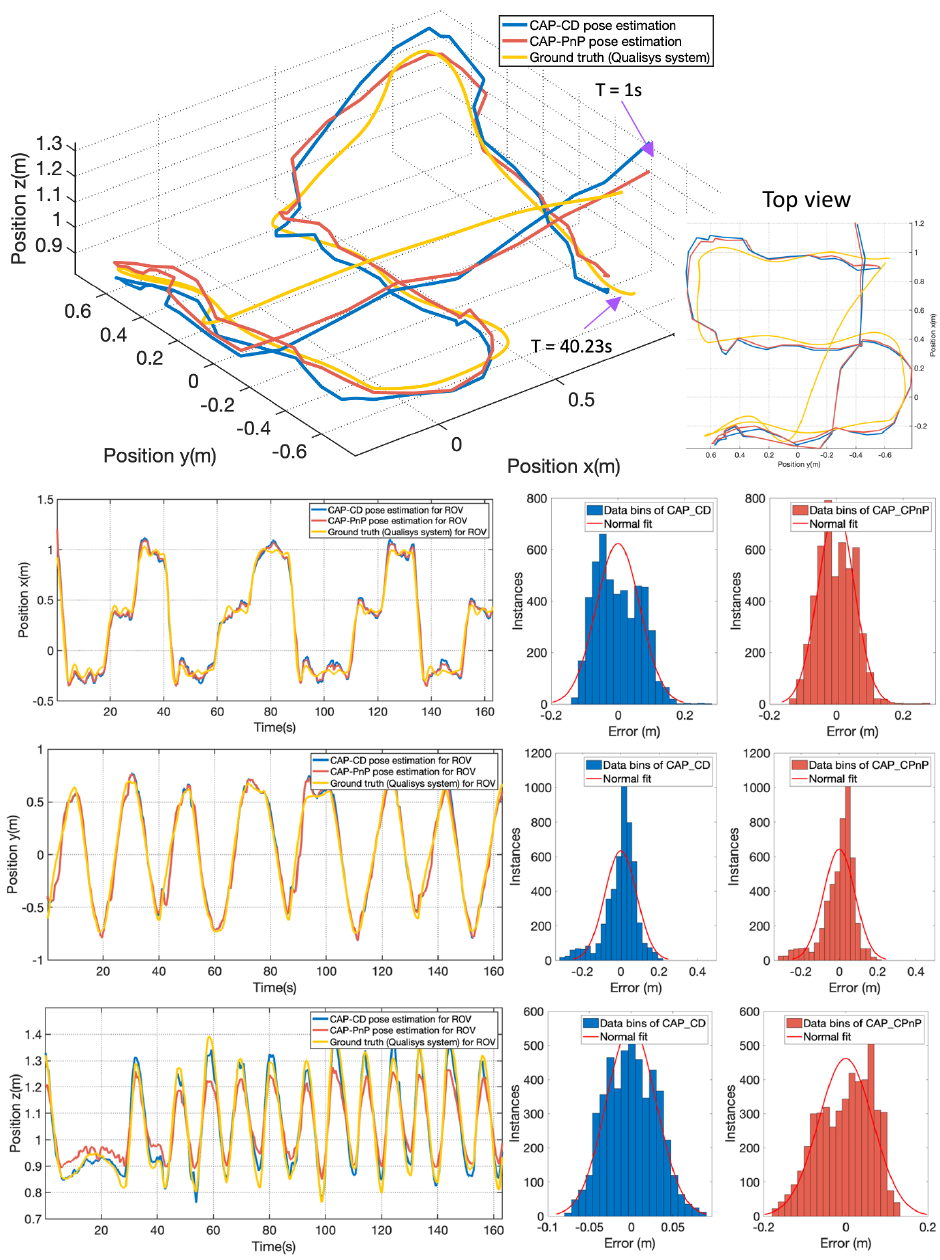}
    \caption{Lawnmower pattern 1}
    \label{fig:lm1}
\end{figure}
\clearpage

\begin{figure}[ht]
    \centering
    \includegraphics[width=0.9\textwidth]{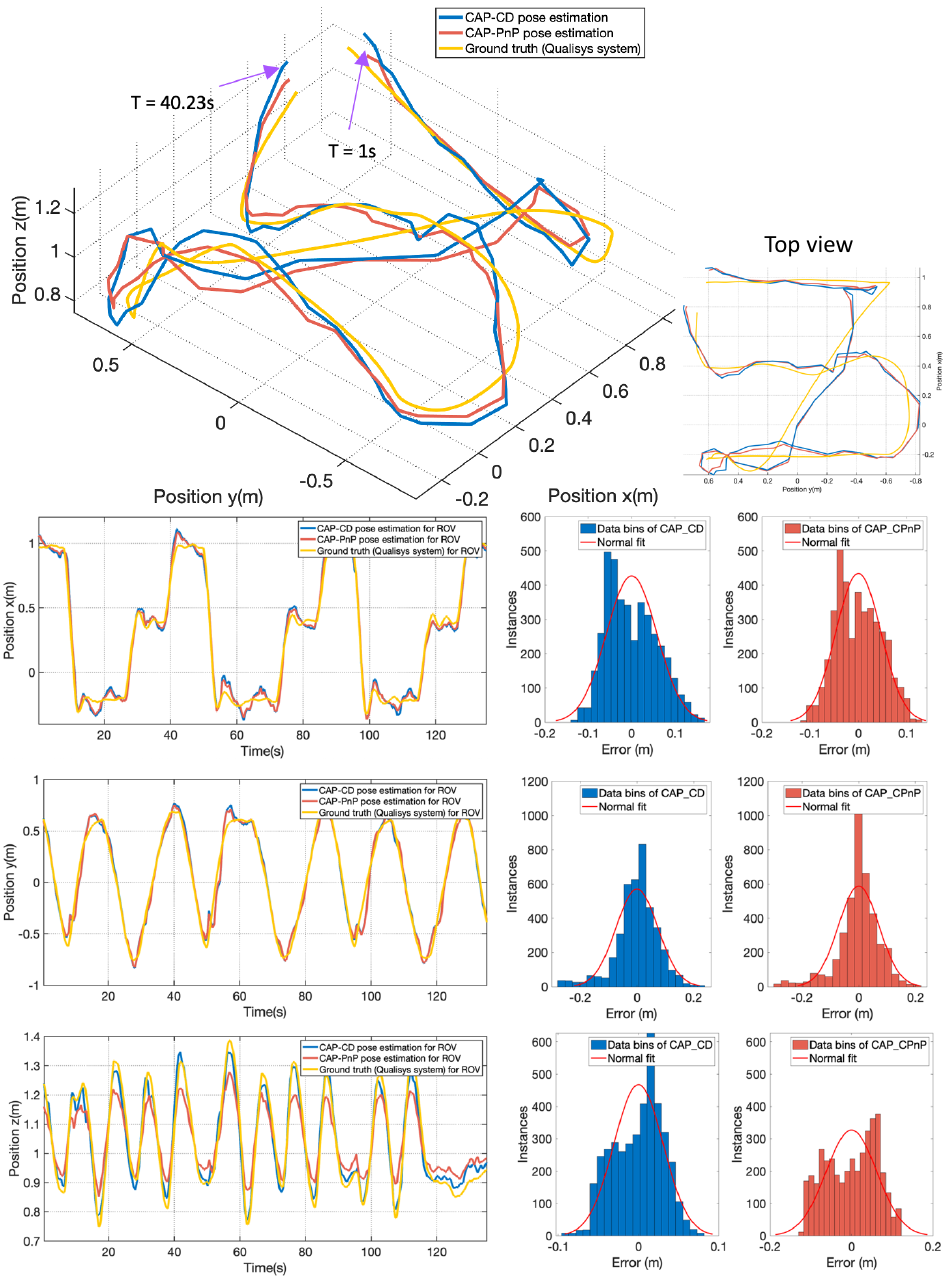}
    \caption{Lawnmower pattern 2}
    \label{fig:lm2}
\end{figure}
\clearpage

\begin{figure}[ht]
    \centering
    \includegraphics[width=0.9\textwidth]{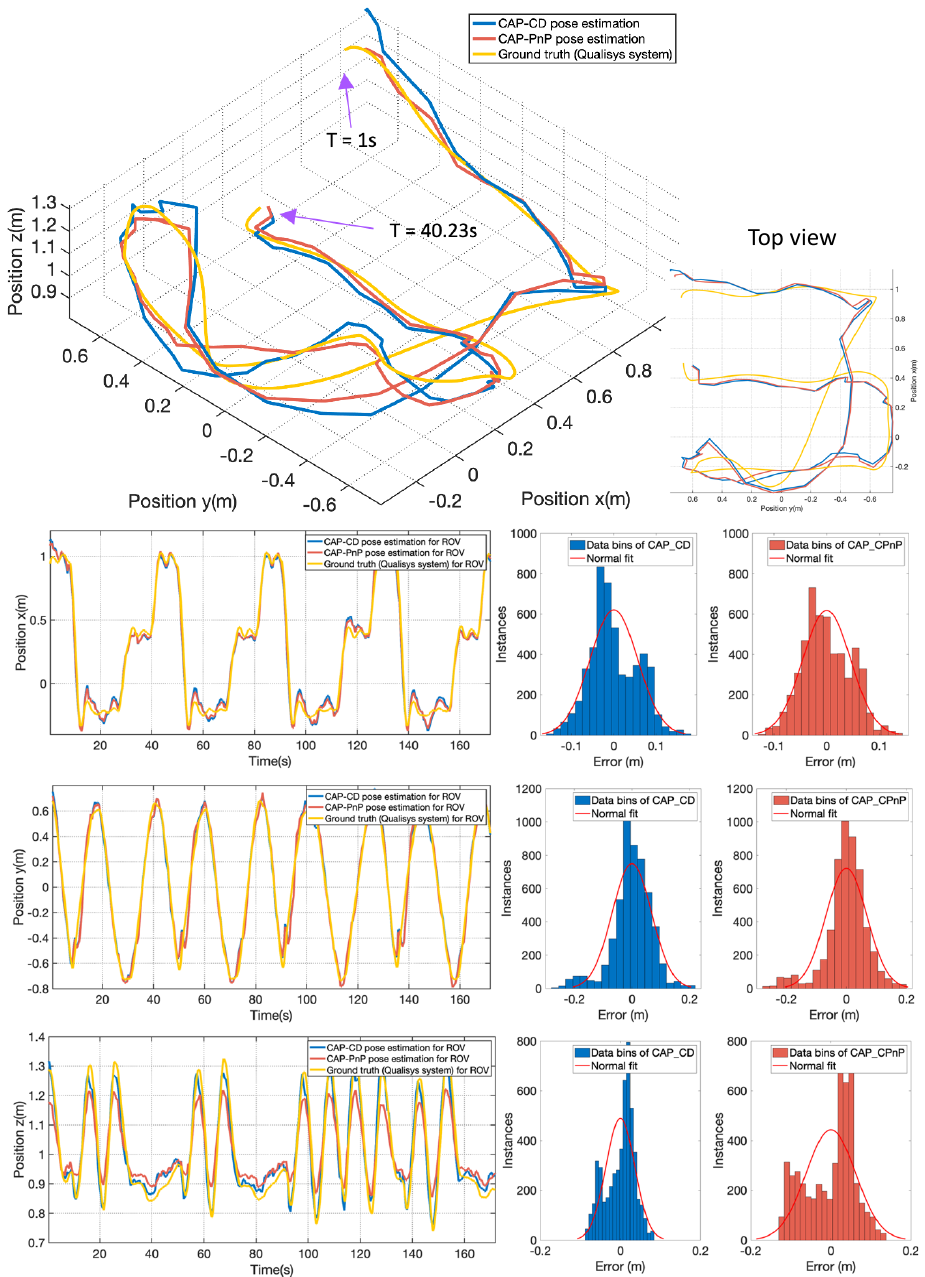}
    \caption{Lawnmower pattern 3}
    \label{fig:lm3}
\end{figure}
\clearpage

\begin{figure}[ht]
    \centering
    \includegraphics[width=0.9\textwidth]{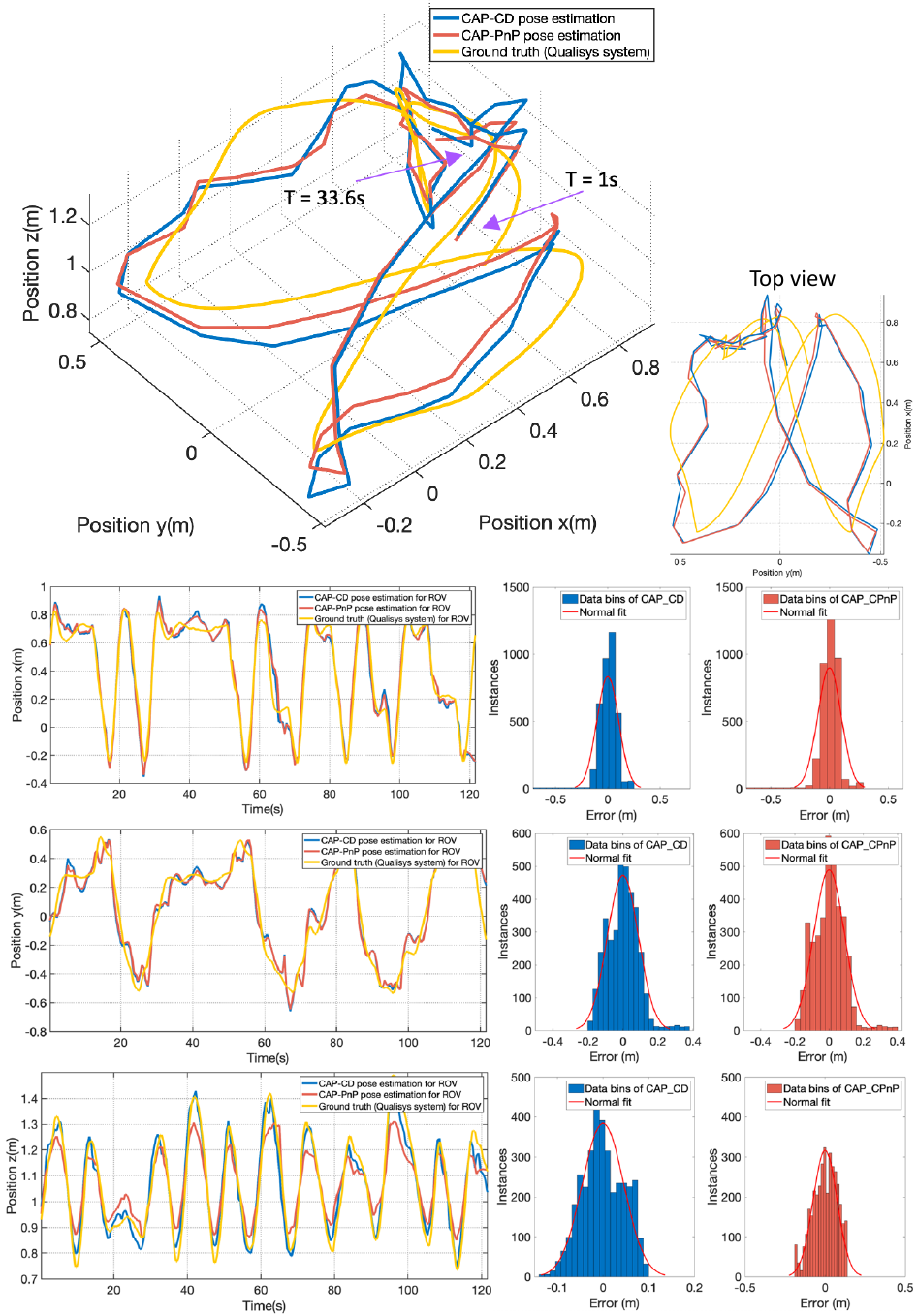}
    \caption{Random pattern 1}
    \label{fig:rd1}
\end{figure}
\clearpage

\begin{figure}[ht]
    \centering
    \includegraphics[width=0.9\textwidth]{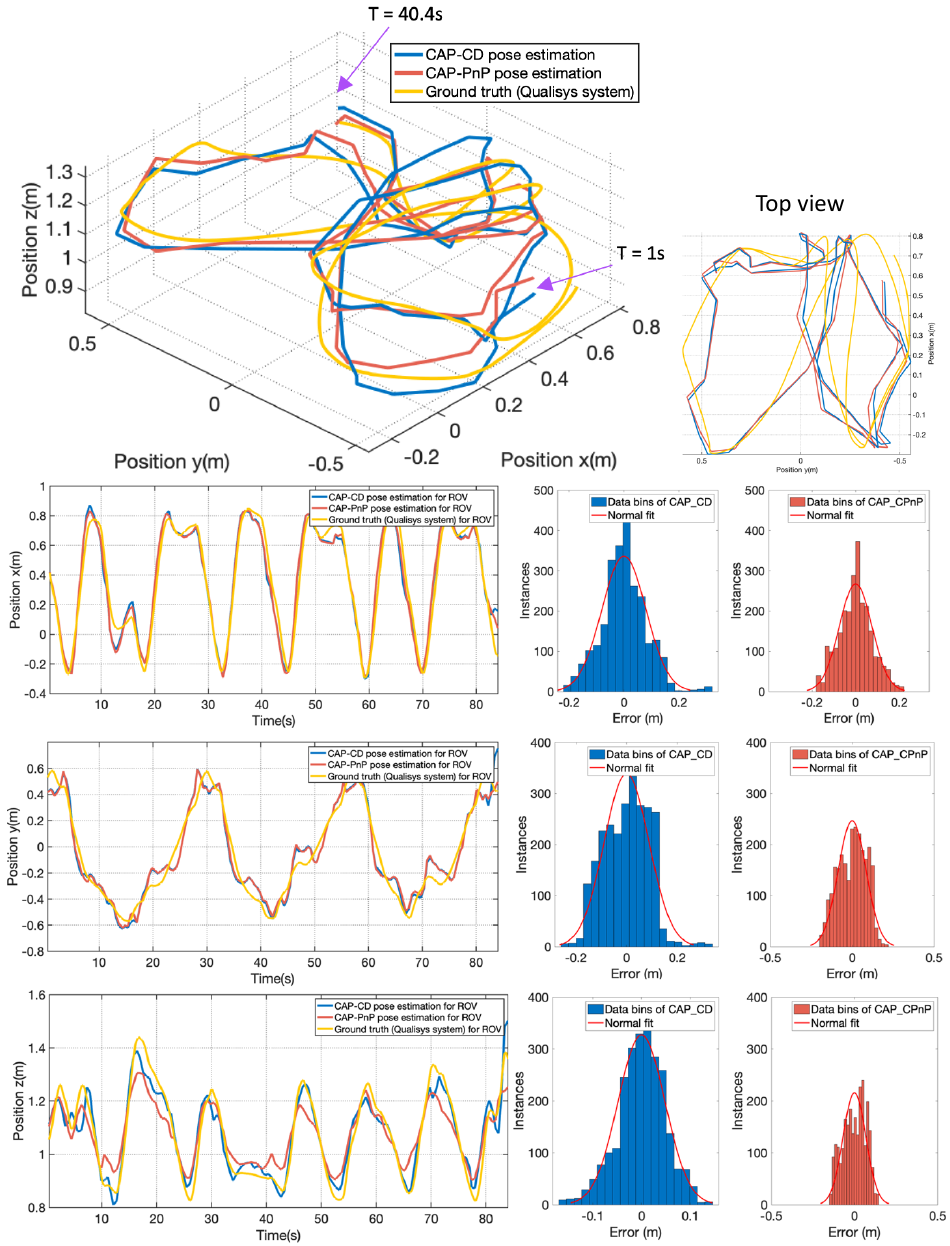}
    \caption{Random pattern 2}
    \label{fig:rd2}
\end{figure}
\clearpage

\begin{figure}[ht]
    \centering
    \includegraphics[width=0.9\textwidth]{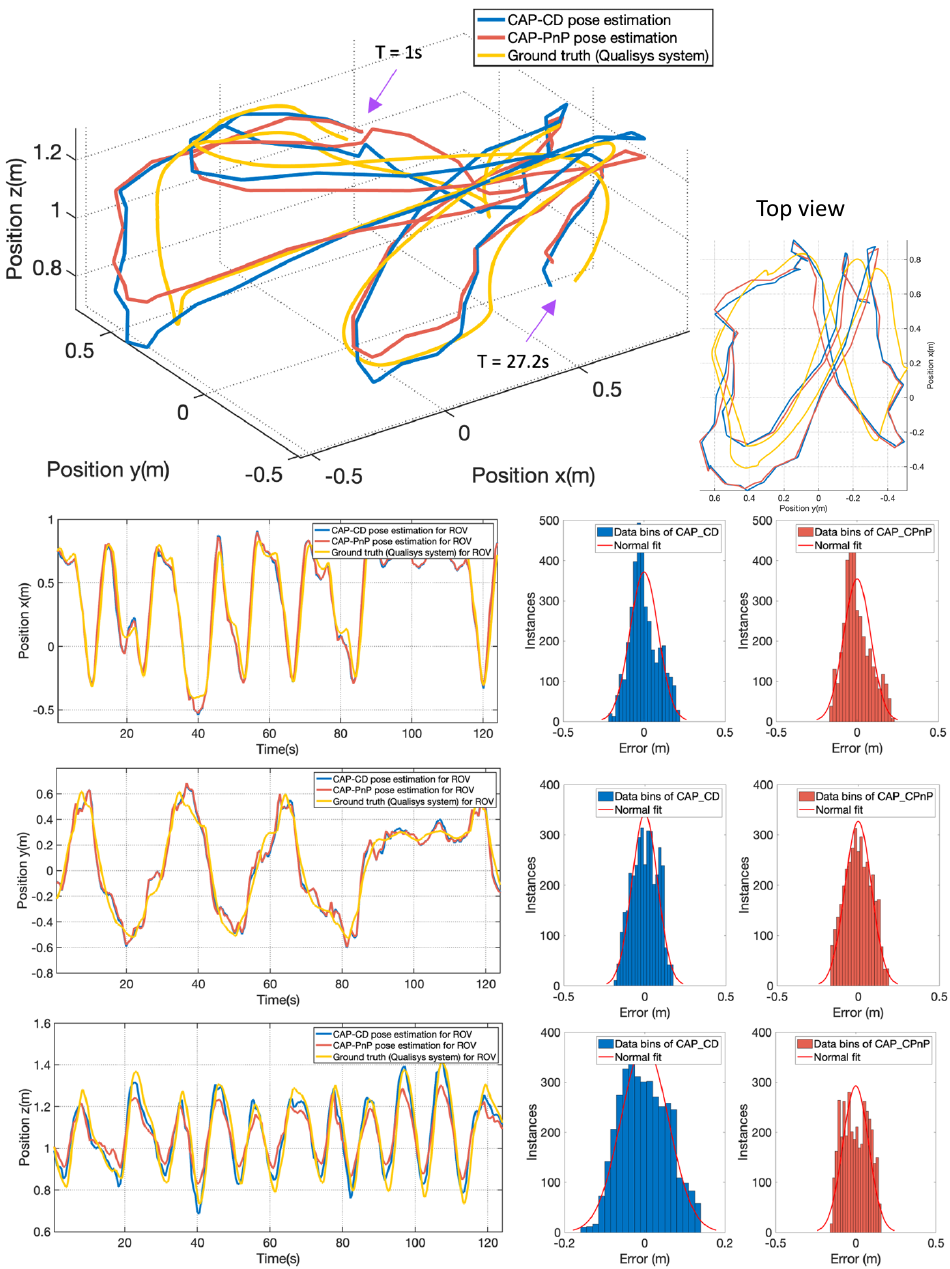}
    \caption{Random pattern 3}
    \label{fig:rd3}
\end{figure}
\clearpage

% \section{Qualisys System Calibration}

% \begin{figure}[ht]
%     \centering
%     \includegraphics[width=\textwidth]{figures/Qualisys_calibration (copy).png}
%     \caption{Calibration result of each camera in Qualisys system}
%     \label{fig:qualisyscali1}
% \end{figure}
% \clearpage
% \begin{figure}[ht]
%     \centering
%     \includegraphics[width=\textwidth]{figures/Qualisys_cali2 (copy).png}
%     \caption{Calibration result of 6 Miqus cameras in Qualisys system}
%     \label{fig:qualisyscali2}
% \end{figure}
% \clearpage
\end{document}